\documentclass[a4paper,conference]{IEEEtran}
\ifCLASSINFOpdf
   \usepackage[pdftex]{graphicx}
\else
\fi
\usepackage{amsmath,amssymb}
\usepackage[inline]{enumitem}
\usepackage{multirow}
\usepackage[table,xcdraw]{xcolor}
\usepackage{color}
\newtheorem{theorem}{Theorem}[section]
\usepackage[marginal]{footmisc}
\usepackage{subfigure}

\begin{document}
%
% paper title
% Titles are generally capitalized except for words such as a, an, and, as,
% at, but, by, for, in, nor, of, on, or, the, to and up, which are usually
% not capitalized unless they are the first or last word of the title.
% Linebreaks \\ can be used within to get better formatting as desired.
% Do not put math or special symbols in the title.
\title{Incrementally Zero-Shot Detection \\by an Extreme Value Analyzer}

% author names and affiliations
% use a multiple column layout for up to three different
% affiliations

\author{\IEEEauthorblockN{Sixiao Zheng}
\IEEEauthorblockA{Academy for Engineering \& Technology, \\Fudan University\\
Shanghai Engineering Research Center \\of AI\& Robotics, \\
Engineering Research Center of \\AI \& Robotics, Ministry of Education\\
Email: sxzheng18@fudan.edu.cn}
\and
\IEEEauthorblockN{Yanwei Fu}
\IEEEauthorblockA{School of Data Science, \\ MOE Frontiers Center for Brain Science, \\Shanghai Key Lab of Intelligent \\ Information Processing,\\ Fudan University, \\
% Email: yanweifu@fudan.edu.cn
}
\and
\IEEEauthorblockN{Yanxi Hou}
\IEEEauthorblockA{School of Data Science, \\
Fudan University\\
% Email: yxhou@fudan.edu.cn
}
}

% use for special paper notices
%\IEEEspecialpapernotice{(Invited Paper)}

% make the title area
\maketitle

% As a general rule, do not put math, special symbols or citations
% in the abstract
\begin{abstract}
   Human beings not only have the ability to recognize novel unseen classes, but also can incrementally incorporate the new classes to 
   existing knowledge preserved. However, zero-shot learning models assume that all seen classes should be known beforehand, while incremental learning models cannot recognize unseen classes. 
   This paper introduces a novel and challenging task of Incrementally Zero-Shot Detection (IZSD), a practical strategy for both zero-shot learning and class-incremental learning in  real-world object detection. An innovative end-to-end model -- IZSD-EVer was proposed to tackle this task that requires  incrementally detecting new classes and detecting the classes that have never been seen.
   Specifically, we propose a novel extreme value analyzer to detect objects from old seen, new seen, and unseen classes, simultaneously. Additionally and technically, we propose two  innovative losses, \emph{i.e.}, background-foreground mean squared error loss alleviating
the extreme imbalance of the background and foreground of images, and projection distance loss aligning the visual space and semantic spaces of old seen classes.
   Experiments demonstrate the efficacy of our model in detecting objects from both the seen and unseen classes, outperforming the alternative models on Pascal VOC and MSCOCO datasets. \footnote{Supported by NSFC Grant 71991471, and NSFS Grant 20ZR1403900.}
\end{abstract}

% no keywords

% For peer review papers, you can put extra information on the cover
% page as needed:
% \ifCLASSOPTIONpeerreview
% \begin{center} \bfseries EDICS Category: 3-BBND \end{center}
% \fi
%
% For peerreview papers, this IEEEtran command inserts a page break and
% creates the second title. It will be ignored for other modes.
\IEEEpeerreviewmaketitle

\section{Introduction}
% no \IEEEPARstart
This paper studies object detection, which, as one important computer
vision task, has seen unprecedented advances in recent years with
the development of Convolutional Neural Network (CNN). However, most
successful object detection models, to data, are formulated as supervised
learning problems in a batch setting. 
Such detection models have limited
the capability of generalizing to unseen object classes, or being
learned in an incremental setting. Humans, on the other hand, not
only have the ability to recognize novel unseen object classes,
but incrementally incorporate these novel object classes to existing
knowledge preserved as well. For instance, a boy visiting the zoo
will continuously identify and remember many new animals whilst not
forget his pet at home.

Previous endeavors tackle object detection in either zero-shot learning (ZSL)
\cite{rahman2018zero}, or class-incremental manner~\cite{shmelkov2017incremental,hao2019ciod}. 
Particularly, model learns to recognize and localize objects from unseen classes in Zero-Shot Detection (ZSD) \cite{rahman2018zero,zhu2019zero,bansal2018zero,rahman2018polarity,rahman2019transductive}.
A successful ZSL model should have effective\emph{ semantic
knowledge transfer}, \emph{i.e.}, transferring knowledge of seen classes,
such as semantic attributes or word vector, to unseen classes. However,
the ZSL assumes that all seen classes should be known beforehand,
and the recognition model is trained in a batch; this greatly and
unrealistically simplified the problem in the real-world setting.
In contrast, the class-incremental learning (CIL) task ~\cite{li2017learning,rebuffi_icarl:_2017,kirkpatrick2017overcoming}
aims at continuously learning newly observed classes without forgetting
existing  old classes, \emph{i.e.}, \emph{catastrophic forgetting}~\cite{goodfellow2014empirical}.
The CIL model is trained via a stream of data in which examples of
different classes occur at different times. Notably, without  semantic
knowledge transfer as ZSL, CIL classifiers are not capable of inferring
of objects of unseen classes in testing.

\begin{figure}[t]
    \centering
    \includegraphics[width=0.8\linewidth]{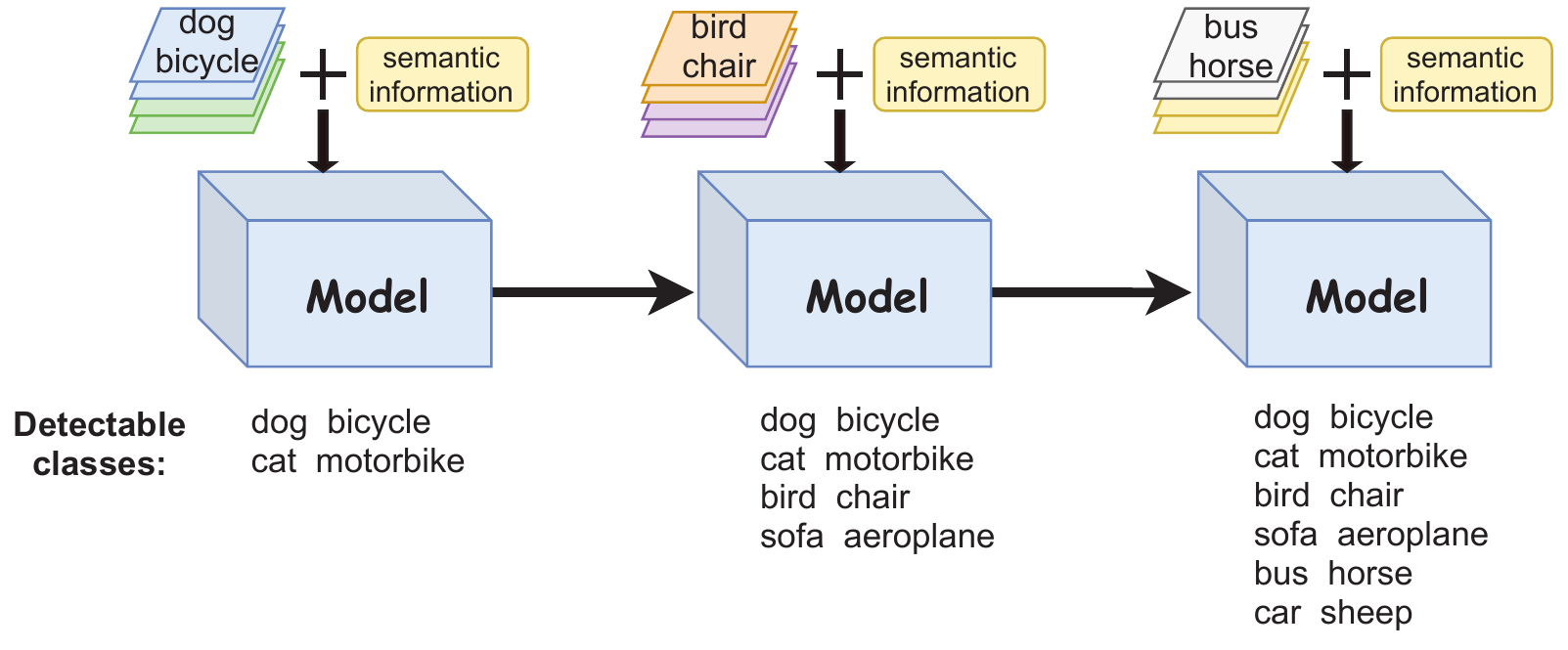}
 \caption{\textbf{Incrementally zero-shot detection.} At each incremental step,
examples of new seen object classes and semantic information are utilized to update the detection model. The model is able to detect objects from old seen, new seen,
and unseen classes. If there is only one incremental step, then IZSD is equivalent to \textbf{zero-shot detection}. In addition, if semantic information is not utilized to detect unseen object classes, the model is equivalent to perform \textbf{class-incremental detection}.}
\label{fig:new_problem}
\end{figure}

Critically, it would be desirable for vision systems to be able to
perform incrementally zero-shot detection (IZSD). Particularly, referring to the properties of CIL 
in iCaRL~\cite{rebuffi_icarl:_2017}, we quantify several properties
of an algorithm qualified as IZSD: (1) an object detection model is
trained from a stream of data, in which visual examples and semantic
information of different object classes are observed at different
incremental steps; (2) the model, at current incremental step, should
not be updated by examples, from existing (\emph{old seen}) object classes,
but only from newly observed (\emph{new seen}) object classes, which
are unseen classes in previous incremental steps, as illustrated in Fig.~\ref{fig:new_problem}.
(3) The trained model should at any time provide a competitive detector
for new seen, old seen and unseen object classes, (4) by the bounded,
memory footprint, and computational cost, with respect to the number
of seen object classes at each incremental step.

These criteria, essentially, identifies the key difference between
IZSD and CIL/ZSL, as well as excluding the trivial solutions in memorizing
all seen objects to retrain a zero-shot detector at each incremental step
(the Fourth Criterion). Specifically, the IZSD should, in principle,
maintain an updated incremental object detector by leveraging both
visual and semantic representations of new seen object classes at
each incremental step. In contrast, the naive solutions by either
visual or semantic cues may be tempted to the deteriorated detection
results. For example, one can directly incrementally update ZSD models  by
using the semantic information from new seen object classes. However,  the predictions of 
such models may be biased towards the old seen object classes, as the corresponding
visual feature extractor is not updated by the visual information from these new seen object classes.
On the other hand, the typical class-incremental object detector \cite{shmelkov2017incremental,hao2019ciod},
updated by the visual cues (but not semantics) of new seen object
classes, is not capable of localizing objects from those classes that
are unseen at the current incremental step. These two naive baselines are compared in Tab. \ref{tab:baselineresult}.

A good IZSD model should be both efficient in semantic knowledge transfer,
and robust to catastrophic forgetting. However, at one incremental
step, the model may be prone to overfitting new seen classes, as they are blind to semantic embeddings of unseen object classes \cite{gzsl_cabliration},
and catastrophically forget the visual representations of old seen
classes. Towards this technical difficulty of IZSD, we propose a novel
extreme value analyzer (EVer), in differentiating the unseen, from
both new and old seen classes by comparing the extreme values~\cite{coles2001introduction}
of each class in the semantic spaces. Furthermore, derived from knowledge
distillation, a novel old-new model is presented to help remember
both the visual and semantic information of old seen classes.

Formally, we propose a novel end-to-end object detection
model for IZSD by EVer
(IZSD-EVer), with the key components of the old-new model and EVer. Specifically, with the backbone of Faster RCNN (FRCN)~\cite{ren2016faster}, the
old-new model maintains the visual and semantic knowledge of old seen
classes, by knowledge distillation~\cite{hinton2015distilling} and projection distance loss, with
a memory component introduced. Furthermore, inspired by 
\emph{Extreme Value Theory} (EVT)~\cite{coles2001introduction},
EVer, in the semantic space, measures the similarity of each instance
to the mean vectors of instances from each seen class by \emph{generalized
Pareto distribution} (GPD)~\cite{pickands1975statistical}.
EVer, in the testing, infers the GPD probability of belonging to
seen or unseen classes, one sample which is further predicted by
either the incremental classifier or zero-shot classifiers. On the benchmark
dataset, we extensively evaluate IZSD-EVer, and the results demonstrate
the efficacy of IZSD-EVer in solving the IZSD task. 

We make four main contributions in this paper: (1) We introduce a
new object detection task of incremental zero-shot detection,
a practical strategy for both ZSL and CIL in the real-world object detection. (2) We propose an innovative
end-to-end trainable model -- IZSD-EVer in addressing IZSD by integrating
ZSL and CIL in a single framework. (3) A novel extreme value analyzer,
is presented, to simultaneously detect objects from old seen, new
seen, and unseen classes. To the best of our knowledge, the GPD in
EVT, is for the first time, introduced for both object detection
and class-incremental learning. (4) Additionally and technically,
we propose two novel losses, \emph{i.e.}, backgroud-forground mean
squared error (bfMSE) alleviating the extreme imbalance of the background
and foreground of images, and projection distance (PD) loss aligning
the visual  and semantic spaces of old seen classes.

% \hfill mds

% \hfill August 26, 2015

\section{Related Works}
\noindent \textbf{Zero-Shot Detection}
The task of ZSL is seeking to recognize unseen visual  categories~\cite{lampert2014attribute,Xian_2016_CVPR,lei2015predicting,zhang2016zero,chao2016empirical,xian2017zero}. As an extension, ZSD aims at  recognizing and localizing the objects of unseen classes in testing~\cite{bansal2018zero,rahman2018zero,zhu2019zero,Demirel2018ZeroShotOD,rahman2018polarity,rahman2019transductive}. Specifically, they adapt the existing detectors (2-stage for~\cite{bansal2018zero,rahman2018zero,rahman2018polarity,rahman2019transductive}, 1-stage for~\cite{zhu2019zero,Demirel2018ZeroShotOD}) in the ZSD setting by adding a separate semantic prediction training task on the annotated seen objects, such that domain knowledge can be transferred to detect unseen objects which are semantically similar to seen objects. Among them, Rahman~\textit{et al.}~\cite{rahman2018polarity} proposed polarity loss explicitly maximizing the margin between predictions for positive and negative classes based on focal loss. In contrast, our IZSD-EVer employs the proposed bgMSE loss as well as reconstruction loss and triplet loss. Generally, the ZSL assumes the model trained in a batch, whilst the proposed IZSD task requires model learning in a class-incremental manner, which is more realistic in real-world applications.

\noindent  \textbf{Class-Incremental Detection}
 Researchers have proposed many different approaches to solve class-incremental learning \cite{kirkpatrick2017overcoming,li2017learning,rebuffi_icarl:_2017,castro2018end,Belouadah_2019_ICCV,Wu_2019_CVPR}. However, these methods focus on solving image classification task rather than more difficult object detection task.
Class-incremental detection aims at adding new classes to well-trained object detector incrementally, using only the data of the new classes, independent with totally re-training. 
Shmelkov \textit{et al.}~\cite{shmelkov2017incremental} presented the first approach to solve the task of class-incremental detection, by a two-stage object detection model on Fast R-CNN \cite{girshick2015fast} with proposals generated by EdgeBoxes \cite{zitnick2014edge}. The model alleviates catastrophic forgetting by optimizing cross-entropy loss and knowledge distillation loss \cite{hinton2015distilling}. 
However, this model is not an end-to-end  model and requires the external EdgeBoxes to generate proposals, which consume many computing resources.
In contrast, Hao \textit{et al.}~\cite{hao2019ciod} proposed an end-to-end architecture for class incremental detection. Based on FRCN, they introduced knowledge distillation in both Region Proposal Network (RPN) and FRCN, transferring knowledge from teacher subnetwork to student subnetwork to generate high-quality box proposals, preventing catastrophic forgetting. Again,  object detection in this setting, is different from our IZSD, as their incapable of predicting objects from unseen classes, and functionally, our EVer is proposed for such a purpose, inspired by recent works on open set 
recognition by  extreme values in computer vision \cite{Scheirer_2011_TPAMI,Scheirer_2014_TPAMIb,Scheirer_2017_MC,Rudd_2018_TPAMI,vignotto2018extreme}.

%------------------------------------------------------------------------
\section{Proposed Model\label{sec:3.1}}

%-------------------------------------------------------------------------

\noindent \textbf{Problem Setup}. In an incremental step, a set of
\emph{old seen} classes denoted as $\mathcal{O}$, whose samples have
been used to train the model in the previous incremental steps. Moreover, 
a set of \emph{new seen} classes $\mathcal{N}$, whose samples are
training data of the current incremental step. There is also a set
of \emph{unseen} classes $\mathcal{U}$, whose samples are only accessible
during the test phase. These three sets of classes are mutually exclusive,
\textit{i.e.}, $\mathcal{O}\cap\mathcal{N}\cap\mathcal{U}=\varnothing$,
so the set of all classes is denoted by $\mathcal{C}=\mathcal{O}\cup\mathcal{N}\cup\mathcal{U}$.
The set of seen classes is $\mathcal{S}=\mathcal{O}\cup\mathcal{N}$.
Besides, $N_{\mathcal{O}}$, $N_{\mathcal{N}}$, $N_{\mathcal{U}}$,
$N_{\mathcal{S}}$, $N_{\mathcal{C}}$ represent the number of old
seen, new seen, unseen, all seen, and all classes, respectively. We
denote semantic information as $\mathbf{E}\in\mathbb{R}^{(C+1)\times d}$
composed of $(C+1)$ semantic embeddings (\textit{e.g.}, word2vec~\cite{mikolov2013distributed}, GloVe \cite{pennington2014glove}, attributes) of $d$ dimensions
as its rows. Each object class has a corresponding semantic embedding,
including the background class. The semantic embedding of the background
class is the mean semantic embedding of other classes, \textit{i.e.},
$\mathbf{E}_{0,:}=\frac{1}{C}\sum_{i=1}^{C}\mathbf{E}_{i,:}$, 
which is the same as \cite{rahman2018zero}. 
Given a model well-trained on old classes dataset
$\mathcal{X_O}$, the IZSD model is
updated by new classes dataset $\mathcal{X_N}$ at current incremental step,
and is capable of generalizing to predict examples from $\mathcal{O}$,
$\mathcal{N}$ and $\mathcal{U}$, individually. Additionally, we
introduce a bounded memory $\mathcal{M}=\{x_{i},i=1,2,\dots,K\}$,
with memory size $K$ is much smaller than the number of $\mathcal{X_{O}}$,
\textit{i.e.}, ${K}\ll N_{\mathcal{O}}$.

\noindent \textbf{End-to-End Learning of IZSD-Ever.} The model is learned in
an end-to-end way. In the first incremental step, since no existing old classes, the new model performs end-to-end learning on $\mathcal{X_{N}}$
through the following loss function. 
\begin{equation}
\mathcal{L}=\mathcal{L}_{bone}+\mathcal{L}_{cls}. \label{equ:Lbone}
\end{equation}
Then we estimate the parameters of GPD by maximum likelihood estimation (MLE) based on the projected
semantic vectors for each new class. We store the
most representative images for each new class in the memory component,
as described in Sec. \ref{sec:ILGZSD}.
In the subsequent incremental steps, the new class dataset $\mathcal{X_{N}}$
and memory $\mathcal{M}$ are simultaneously inputted into the old 
and new models, and the new model is trained end-to-end using
the following loss function. 
\begin{equation}
\mathcal{L}=\mathcal{L}_{bone}+\mathcal{L}_{IL}.
\end{equation}
For the new classes, we estimate the parameters of the GPD for them
as in the first incremental step. The parameters of GPD of the old
classes remain unchanged. The memory $\mathcal{M}$ is also updated accordingly.
We will introduce each loss in the following subsection.

\subsection{Backbone with Semantic Embedding }

\begin{figure*}[t]
\centering 
\includegraphics[width=0.75\linewidth]{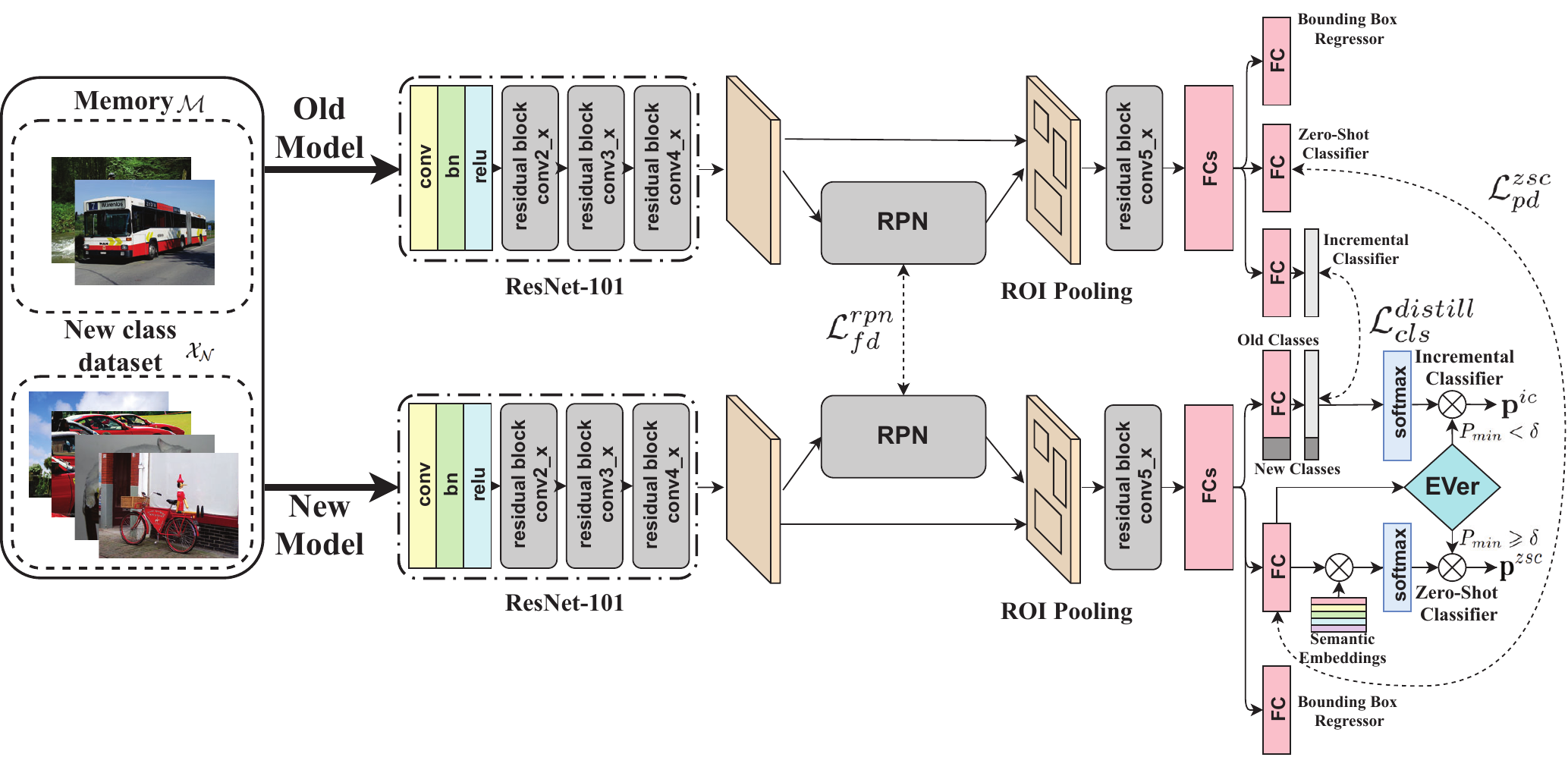} 
 \caption{The overall architecture of our IZSD-EVer model. \label{fig:framework} }

\end{figure*}

Built upon the backbone of FRCN, we additionally add
the semantic embedding. Particularly, as shown in Fig. \ref{fig:framework},
we denote the visual feature of the fully connected (FC) layer
before zero-shot classifier as $f(x;\Theta)\in\mathbb{R}^{v}$, where
$x$ and $\Theta$ denote the input image, the weights of FRCN
model, individually. It is followed by another FC layer for zero-shot classifier,
projecting visual features from visual space to $d$-dimensional semantic
space denoted by $\mathbf{W}\in\mathbb{R}^{d\times v}$, as $\mathbf{s}=\mathbf{W}f(x;\Theta),\quad\mathbf{s}\in\mathbb{R}^{d}$.
Cosine similarity is employed to measure the distance between projected
semantic vector $\mathbf{s}$ and semantic embedding of classes $\mathbf{E}$,
and output the predicted class probability, as $\mathbf{p}^{zsc}=\mathtt{softmax}(\mathbf{Es}),\quad\mathbf{p}^{zsc}\in\mathbb{R}^{N_{\mathcal{C}}+1}$,
where $\mathbf{s}$ and $\mathbf{E}$ are $\ell^{2}$ normalized.

To efficiently learn the trainable $\mathbf{W}$, we present
three loss functions to help align  visual features and with corresponding semantic embeddings.
Particularly, 

\noindent \textbf{bfMSE Loss}. We train $\mathbf{W}$ using background-foreground
MSE (bfMSE) loss to alleviate extreme imbalance between background and foreground
in proposals generated by RPN. We separately compute 
MSE loss between the projected semantic vectors of background and
foreground and the corresponding semantic embeddings. The bfMSE loss
is defined as
\begin{equation}
    \begin{aligned}
        \mathcal{L}_{bfmse} &=\frac{1}{N_{bg}}\sum_{i=1}^{N_{bg}}\mathbb{I}_{i}^{bg}\|\mathbf{s}_{i}-\mathbf{E}_{0,:}\|_{2}^{2} \\
        &+\alpha\frac{1}{N_{obj}}\sum_{i=1}^{N_{obj}}\mathbb{I}_{i}^{obj}\|\mathbf{s}_{i}-\mathbf{E}_{y_{i},:}\|_{2}^{2},
    \end{aligned}
    \label{equ:mse}
\end{equation}

\noindent where $\mathbb{I}_{i}^{bg}$ and $\mathbb{I}_{i}^{obj}$ are indicators
showing that the $i$-th region proposal contains background and object,
respectively. $N_{bg}$ and $N_{obj}$ are the number of region proposals
belonging to the background and foreground, respectively. $y_{i}$
is the ground-truth label of $i$-th region proposal. We empirically
set $\alpha=5$.

\noindent \textbf{Reconstruction Loss}. We introduce this loss in \cite{kodirov2017semantic}
to learn a more generalized model, as 
\begin{equation}
\mathcal{L}_{rec}=\frac{1}{N_{p}}\sum_{i=1}^{N_{p}}\|f(x;\Theta)-\mathbf{W}^{T}\mathbf{s}\|_{2}^{2},\label{equ:prb}
\end{equation}
\noindent where $\mathbf{W}^{T}$ project $\mathbf{s}$ back to visual space
to reconstruct the visual feature. $N_{p}=N_{bg}+N_{obj}$ is the
number of proposals. Then we compute the MSE loss between the reconstructed
visual features and the original visual features. Note that $\mathcal{L}_{bfmse}$
and $\mathcal{L}_{rec}$ are to constrain the projected semantic vectors
to be similar to semantic embedding. 

\noindent \textbf{Triplet Loss}. This loss enforces the projected
semantic vectors are more similar semantic embedding of ground-truth
classes, than those from other classes, defined as
\begin{equation}
\mathcal{L}_{tri}=\frac{1}{N_{p}}\sum_{i=1}^{N_{p}}\sum_{\substack{j=1\\
j\neq y_{i}}}^{N_{\mathcal{S}}}\max(m+C_{ij}-C_{iy_{i}},0)\label{equ:tri},
\end{equation}
\noindent where $C_{ij}$ is the cosine similarity between the $\mathbf{s}_{i}$
and $\mathbf{E}_{j,:}$. Similarly, $C_{iy_{i}}$ is the cosine similarity
between the $\mathbf{s}_{i}$ and its ground-truth semantic embedding
$\mathbf{E}_{y_{i},:}$. In addition, the standard losses in FRCN
are also utilized here as $\mathcal{L}_{FRCN}=\mathcal{L}_{cls}^{rpn}+\mathcal{L}_{reg}^{rpn}+\mathcal{L}_{reg}$,
\noindent where $\mathcal{L}_{rpn}^{cls}$, $\mathcal{L}_{rpn}^{reg}$ and $\mathcal{L}_{reg}$
are the classification loss in RPN, bounding box regression loss in
RPN and bounding box regression loss of bounding box regressor, respectively.

\begin{equation}
    \mathcal{L}_{FRCN}=\mathcal{L}_{cls}^{rpn}+\mathcal{L}_{reg}^{rpn}+\mathcal{L}_{reg}
\end{equation}

The overall loss function of the backbone model is 
\begin{align}
\mathcal{L}_{bone} & =\mathcal{L}_{FRCN}+\mathcal{L}_{bfmse}+\beta\mathcal{L}_{rec}+\mathcal{L}_{tri},\label{equ:lzsd}
\end{align}
where $\beta$ is a hyperparameter to control the importance of different losses.

\subsection{The Old-New Model\label{sec:ILGZSD}}

The bounded memory component $\mathcal{M}$ \cite{castro2018end,Wu_2019_CVPR,Belouadah_2019_ICCV,rebuffi_icarl:_2017}
is introduced into our old-new model to alleviate the imbalance between
old seen and new seen classes. 
Particularly, $\mathcal{M}$ stores a few images for each seen classes based on their representative object example visual features $f(x;\Theta)$, which closest to the mean feature vector of the classes.
Since $\mathcal{M}$ has a limited memory capacity of $K$, it will clear the least representative images of each class as new seen classes come.

The old-new model is presented to incrementally learn new seen classes
without catastrophic forgetting old seen classes. Specifically, as
in Fig.~\ref{fig:framework}, we add an incremental classifier
to the backbone model. When new seen classes $\mathcal{N}$ occurs at the current
incremental step, we combine the \textit{old} model learned in previous steps,
with the \textit{new} model learned in this step, forming the old-new model.
When incremental learning is performed, new class dataset  $\mathcal{X}_{\mathcal{N}}$ and memory $\mathcal{M}$ are simultaneously entered into the old-new model, where the parameters of the old model are fixed. 
 When our old-new model learns new classes, the RPN of the new model
needs to be able to generate region proposals containing objects of
the old or new classes. We re-purpose the domain expansion in RPN
of \cite{hao2019ciod} to maintain the knowledge of the old class.
Thus, we freeze the RPN classifier to keep the decision boundary of
the RPN classifier unchanged and update the parameters of other parts
of the RPN,  making the  feature distance loss. 
\begin{equation}
\mathcal{L}_{fd}^{rpn}=\|f_{rpn}(x;\Theta_{o})-f_{rpn}(x;\Theta_{n})\|_{F}^{2}\label{equ:Ldisrpn}.
\end{equation}
$f_{rpn}(\cdot)$ denotes the feature map generated by the intermediate
convolutional layer of RPN. This loss ensures that the RPN generates proposals containing objects of new or old classes.

\begin{figure}[t]
\centering 
\includegraphics[width=0.8\linewidth]{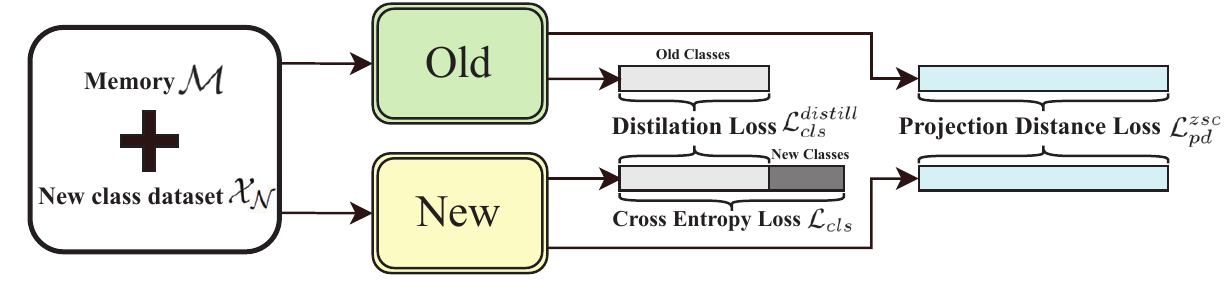} 
 \caption{Diagram of incremental learning for IZSD.}
\label{fig:distill} 
\end{figure}

When learning with new classes, the number of output nodes of the incremental classifier in the new model is expanded to the number of classes that have seen.
As shown in Fig.~\ref{fig:distill}, besides the traditional cross-entropy loss on
all classes, we add a knowledge distillation loss~\cite{hinton2015distilling} on the incremental classifier 
to transfer knowledge from the old model to the new model to alleviate
performance dropping on old seen classes. Besides, we also add a projection distance (PD) loss on the zero-shot classifier to maintain the ability to align the visual space and semantic spaces of old seen classes.
Particularly, let us denote
the projected semantic embeddings of the old and new models as $\mathbf{s}_{o}$
and $\mathbf{s}_{n}$, respectively. Similarly,
we denote the output logits of the incremental classifier as $\hat{y}_{o}$
and $\hat{y}_{n}$ for the old and new models, respectively. 
\begin{align}
\mathcal{L}_{cls}^{distill} & =\mathtt{CE}(\mathtt{softmax}(\frac{\hat{y}_{o}}{T}),\mathtt{softmax}(\frac{\hat{y}_{n}^{o}}{T})),\\
\mathcal{L}_{cls} & =\mathtt{CE}(y,\mathtt{softmax}(\hat{y}_{n})),\\
\mathcal{L}_{pd}^{zsc} & =\|\mathbf{s}_{o}-\mathbf{s}_{n}\|_{2}^{2}, \label{equ:Lpd}
\end{align}
\label{equ:inc}
\noindent where $\hat{y}_{n}^{o}$ is the output logits of the new
model for old classes. $y$ is the ground-truth label. $T$ is
a temperature scalar. $\mathtt{CE}(\cdot)$ represents the cross-entropy
loss. The overall loss function for incremental learning as follows:
\begin{equation}
\mathcal{L}_{IL}=\frac{N_{\mathcal{O}}}{N_{\mathcal{C}}}\left(\mathcal{L}_{cls}^{distill}+\mathcal{L}_{pd}^{zsc}\right)+\frac{N_{\mathcal{N}}}{N_{\mathcal{C}}}\mathcal{L}_{cls}+\gamma\mathcal{L}_{fd}^{rpn}.\label{equ:Lil}
\end{equation}
The first three items are multiplied by the ratio of the class they
are responsible for, indicating their importance during training.
It reduces the hyperparameters and makes the model easier to train. And $\gamma$ is a hyperparameter.

\subsection{Extreme Value Analyzer}

We construct an EVer as the Pickands-Balkema-de Haan Theorem \cite{balkema1974residual}
to differentiate unseen from seen object classes. Typically, let $X_{1},X_{2},\dots,X_{n}$
be a sequence of identically distributed random variables (\emph{i.i.d})
with cumulative distribution function $F$. The conditional excess
distribution function of $X$ over a threshold $u$ is defined as
\begin{equation}
      F_u(x) = \Pr(X-u \leqslant x|X>u) = \frac{F(x+u)-F(u)}{1-F(u)}, x \geqslant 0.
\end{equation}

%------------------------------------------------------------------------

\begin{figure}[t]
    \centering
    \includegraphics[width=0.6\linewidth]{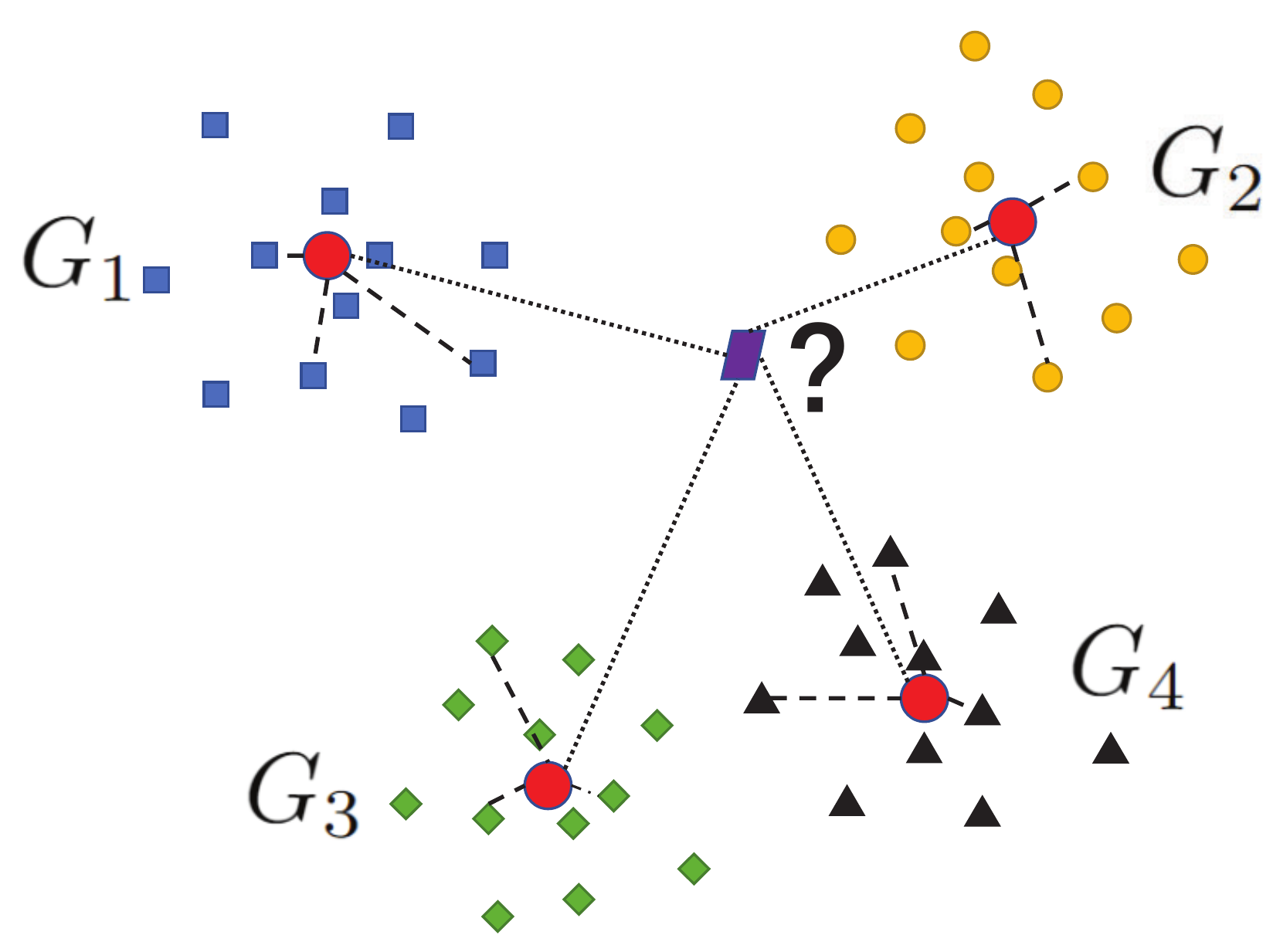}
    \caption{Extreme Value Analyzer. $G_1,G_2,G_3,G_4$ represent the fitted GPD.}
    \label{fig:EVTfitting}
\end{figure}
%------------------------------------------------------------------------

\begin{theorem}[Pickands–Balkema–de Haan theorem] \cite{balkema1974residual}
The conditional excess distribution function $F_u$ can be well approximated by the generalized Pareto distribution when the threshold $u$ is sufficiently large. That is 
\begin{equation}
   F_u(x) \to G(x;\sigma, \xi), \quad \text{as} ~ u \to \infty
\end{equation}
\end{theorem}

The GPD is defined as 
\begin{equation}
   \begin{aligned}
      G(x;\sigma, \xi) = \left\{
         \begin{array}{ll}
            1-\left(1+\xi\frac{x}{\sigma}\right)^{-\frac{1}{\xi}}, & \xi \neq 0, \\
            1-\exp\left(-\frac{x}{\sigma}\right), & \xi = 0,
            \label{equ:GPD}
         \end{array}\right.
   \end{aligned}
\end{equation}
$x \geqslant \mu$  when $\xi \geqslant 0$, $\mu \leqslant x \leqslant \mu-\frac{1}{\xi}$ when $\xi < 0$.  

In the semantic space, the projected semantic vectors of the same
class are close together, and the projected semantic vectors of the
different classes are far away from each other. As shown in Fig. \ref{fig:EVTfitting},
we propose to use the GPD of EVT to model the projected semantic vectors
for each seen class. Particularly, if the distance between one projected semantic vectors to the mean projected semantic vector exceeds a threshold,
such projected semantic vector would be taken as an extreme semantic vector.
To this end, we fit the GPD based on the distance beyond the threshold.

We first compute the Euclidean distance $d_{ij}$ between the projected
semantic vectors $\mathbf{s}_{j}^{i}$ and the mean projected semantic
vector $\overline{\mathbf{s}}_{j}$ of $j$-th class, \textit{i.e.},
$d_{ij}=\|\mathbf{s}_{j}^{i}-\overline{\mathbf{s}}_{j}\|_{2}, j\in\{1,2,\dots,S\}$, where $\mathbf{s}_{j}^{i}$ and $\overline{\mathbf{s}}_{j}$ are $\ell^2$ normalized. Then we get the estimated parameters $\hat{\sigma}_{j}$
and $\hat{\xi}_{j}$ of GPD by MLE
based on the excess of distance exceeding threshold $u_{j}$
for $j$-th class. In classifying, we compute the excess of distance $\|\mathbf{s}-\overline{\mathbf{s}}_{j}\|_{2}$ beyond threshold
$u_{j}$, and then derive the probability that the semantic vector
is an extreme semantic vector by substituting the excess into the
GPD, that is, the probability of not belonging to the $j$-th class.
We get $P_{min}$ over seen classes as follows.
\begin{equation}
P_{min}=\min_{j\in\{1,2,\dots,S\}}G(\|\mathbf{s}-\overline{\mathbf{s}}_{j}\|_{2}-u_{j},\hat{\sigma}_{j},\hat{\xi}_{j}).
\end{equation}

We obtain the prediction class label for the region proposal by the
following 
\begin{equation}
\begin{aligned}\hat{y}=\left\{ \begin{array}{ll}
\mathop{\arg\max}_{i\in\{1,2,\dots,S\}}\mathbf{p}_{i}^{ic}, & P_{min}<\delta,\\
\mathop{\arg\max}_{i\in\{1,2,\dots,U\}}\mathbf{p}_{i}^{zsc}, & P_{min}\geqslant\delta, \label{equ:pred}
\end{array}\right.\end{aligned}
\end{equation}
\noindent where $\delta$ is a threshold. If $P_{min}\geqslant\delta$, then
$\mathbf{s}$ is an extreme semantic vector for all seen classes, the corresponding region proposal belongs to unseen classes. Conversely,
if $P_{min}<\delta$, the corresponding region proposal belongs to
seen classes.

%------------------------------------------------------------------------

\section{Experiments}

\subsection{Datasets and Settings}\label{sec:data_setting}
\noindent\textbf{Dataset:} We evaluate our model on Pascal VOC 2007 and 2012 datasets~\cite{everingham_pascal_2010} and MSCOCO 2014 and 2017 datasets~\cite{lin_microsoft_2014}, respectively. 
MSCOCO is a large-scale dataset for object detection, segmentation, and captioning. This dataset is more challenging than Pascal VOC as it has 80 object classes, more small objects, and more complex background.  Our model is compared in CIL and ZSL settings. Specifically, for class-incremental detection, we use the same datasets VOC 2007 and MSCOCO 2017 as \cite{hao2019ciod,shmelkov2017incremental}. 
For zero-shot detection, we use VOC 2007 and MSCOCO 2014 as \cite{Demirel2018ZeroShotOD,rahman2018polarity}. 
In order to compare fairly with other ZSD methods, we use the train split of VOC 2007 and 2012 (with 20 object classes) during training and use val+test split of VOC 2007 for testing. 

\noindent\textbf{Evaluation Metric:} We use the standard average precision (AP) at 0.5 IoU threshold over each class, and mean average precision (mAP) over all the classes as the evaluation metric.

\begin{table}[t]
   \centering
   \caption{Old classes, new class and unseen classes split in different incremental step.}   
   \label{tab:classsplit}
      \small
      \setlength{\tabcolsep}{0.4em}
      \begin{tabular}{c|l|c|r|c|c}
         \hline
         Step & \multicolumn{1}{c|}{\begin{tabular}[c]{@{}c@{}}Old \\ classes\end{tabular}} & \begin{tabular}[c]{@{}c@{}}New \\ classes\end{tabular} & \multicolumn{1}{c|}{\begin{tabular}[c]{@{}c@{}}Unseen \\ classes\end{tabular}} & Train data                        & Test data                                        \\ \hline
         1  & -                                                                           & $\mathcal{G}_1$                                        & $\mathcal{G}_2,\mathcal{G}_3,\mathcal{G}_4$                                    & $\mathcal{D}_{tr}(\mathcal{G}_1)$ & \multirow{4}{*}{$\mathcal{D}_{te}(\mathcal{C})$} \\ \cline{1-5}
         2  & $\mathcal{G}_1$                                                             & $\mathcal{G}_2$                                        & $\mathcal{G}_3,\mathcal{G}_4$                                                  & $\mathcal{D}_{tr}(\mathcal{G}_2)$ &                                                  \\ \cline{1-5}
         3  & $\mathcal{G}_1,\mathcal{G}_2$                                               & $\mathcal{G}_3$                                        & $\mathcal{G}_4$                                                                & $\mathcal{D}_{tr}(\mathcal{G}_3)$ &                                                  \\ \cline{1-5}
         4  & $\mathcal{G}_1,\mathcal{G}_2,\mathcal{G}_3$                                 & $\mathcal{G}_4$                                        & -                                                                              & $\mathcal{D}_{tr}(\mathcal{G}_4)$ &                                                  \\ \hline
         \end{tabular}
   \end{table}

\begin{table*}[t]
\centering
\caption{The detail results of IZSD-EVer on VOC 2007}
\label{tab:IZSD_VOC}
\small
\setlength{\tabcolsep}{0.25em}
\begin{tabular}{c|cccccccccccccccccccc|c|c|c}
\hline
Step & \rotatebox{90}{\begin{tabular}[c]{@{}l@{}}aero-\\ plane\end{tabular}} & \rotatebox{90}{bicycle} & \rotatebox{90}{bird}  & \rotatebox{90}{boat}  & \rotatebox{90}{bottle} & \rotatebox{90}{bus}   & \rotatebox{90}{car}   & \rotatebox{90}{cat}   & \rotatebox{90}{chair} & \rotatebox{90}{cow}   & \rotatebox{90}{\begin{tabular}[c]{@{}l@{}}dining \\ table\end{tabular}} & \rotatebox{90}{dog}   & \rotatebox{90}{horse} & \rotatebox{90}{\begin{tabular}[c]{@{}l@{}}motro-\\ bike\end{tabular}} & \rotatebox{90}{person} & \rotatebox{90}{\begin{tabular}[c]{@{}l@{}}potted \\ plant\end{tabular}} & \rotatebox{90}{sheep} & \rotatebox{90}{sofa}  & \rotatebox{90}{train} & \rotatebox{90}{\begin{tabular}[c]{@{}l@{}}tvmo-\\ nitor\end{tabular}} & \rotatebox{90}{Seen}  & \rotatebox{90}{Unseen} & \rotatebox{90}{mAP}   \\ \hline
1    & 57.2      & 72.8    & 31.7 & 41.3 & 24.3   & 0.9  & 6.9  & 0.2  & 0.0   & 0.3  & 0.0         & 0.3  & 0.3   & 26.3      & 0.4    & 0.0         & 1.0   & 0.0  & 9.5   & 0.1       & 45.5 & 3.1    & 13.7 \\
2    & 50.3      & 61.3    & 51.3 & 29.8 & 22.2   & 54.0 & 74.9 & 73.6 & 23.1  & 41.5 & 0.1         & 9.1  & 5.8   & 16.8      & 0.8    & 0.1         & 1.1   & 13.9 & 10.9  & 0.2       & 48.2 & 5.9    & 27.0 \\
3    & 51.2      & 54.4    & 51.2 & 31.0 & 26.4   & 45.3 & 72.7 & 66.2 & 22.2  & 37.4 & 62.3        & 64.4 & 73.2  & 69.7      & 67.2   & 0.0         & 3.2   & 12.1 & 12.6  & 0.1       & 53.0 & 5.6    & 41.1 \\
4    & 59.1      & 67.0    & 61.3 & 39.8 & 36.7   & 51.9 & 74.9 & 68.4 & 19.8  & 38.8 & 41.9        & 58.7 & 71.7  & 73.5      & 69.6   & 29.3        & 60.8  & 67.4 & 72.5  & 59.2      & 56.1 & -      & 56.1 \\ \hline
\end{tabular}
\end{table*}

\begin{table*}[t]
\centering
  \caption{The results of class-incremental detection on VOC 2007 and COCO 2017. Gray region means that the results of these groups can only be predicted in our model}
  \label{tab:overallresult}
      \small
      \setlength{\tabcolsep}{0.4em}
\begin{tabular}{c|c|cccc|c|c|cccc|c|c|cccc|c}
\hline
Dataset                                                               & Method                                                                                     & $\mathcal{G}_1$ & $\mathcal{G}_2$    & $\mathcal{G}_3$    & $\mathcal{G}_4$    & mAP  & Method                                                                & $\mathcal{G}_1$    & $\mathcal{G}_2$    & $\mathcal{G}_3$    & $\mathcal{G}_4$    & mAP  & Method                  & $\mathcal{G}_1$                        & $\mathcal{G}_2$                                               & $\mathcal{G}_3$                                                                      & $\mathcal{G}_4$                                                                       & \multicolumn{1}{c}{mAP} \\ \hline
                                                                      &                                                                                            & 61.9            & -    & -    & -    & 61.9 &                                                                       & 63.9 & -    & -    & -    & 63.9 &                         & 66.3                     & -                                               & -                                                                      & -                                                                       & 66.3                     \\
                                                                      &                                                                                            & 33.3            & 67.8 & -    & -    & 50.6 &                                                                       & 43.8 & 71.2 & -    & -    & 57.5 &                         & 44.8                     & 59.2                                            & -                                                                      & -                                                                       & 52.0                     \\
                                                                      &                                                                                            & 13.5            & 23.1 & 55.8 & -    & 30.8 &                                                                       & 35.3 & 49.0 & 68.4 & -    & 50.9 &                         & 49.0                     & 43.4                                            & 48.6                                                                   & -                                                                       & 47.0                     \\
                                                                      & \multirow{-4}{*}{\begin{tabular}[c]{@{}c@{}}Faster\\ RCNN\\ (Fine-\\ tuning)\end{tabular}} & 14.2            & 21.5 & 44.0 & 52.5 & 33.1 & \multirow{-4}{*}{CIFRCN}                                              & 34.6 & 44.1 & 55.6 & 59.6 & 48.5 & \multirow{-4}{*}{IDWCF} & 40.2                     & 35.5                                            & 42.5                                                                   & 38.8                                                                    & 39.3                     \\ \cline{2-19} 
                                                                      &                                                                                            & 64.0            & -    & -    & -    & 64.0 &                                                                       & 56.8 & -    & -    & -    & 56.8 &                         & \multicolumn{1}{l}{45.5} & \multicolumn{1}{l}{\cellcolor[HTML]{C0C0C0}1.7} & \multicolumn{1}{l}{\cellcolor[HTML]{C0C0C0}5.5}                        & \multicolumn{1}{l|}{\cellcolor[HTML]{C0C0C0}2.1}                        & \multicolumn{1}{l}{13.7} \\
                                                                      &                                                                                            & 65.2            & 73.2 & -    & -    & 69.2 &                                                                       & 40.4 & 68.8 & -    & -    & 54.6 &                         & \multicolumn{1}{l}{43.0} & \multicolumn{1}{l}{53.4}                        & \multicolumn{1}{l}{\cellcolor[HTML]{C0C0C0}{\color[HTML]{000000} 6.5}} & \multicolumn{1}{l|}{\cellcolor[HTML]{C0C0C0}{\color[HTML]{000000} 5.2}} & \multicolumn{1}{l}{27.0} \\
                                                                      &                                                                                            & 66.2            & 71.3 & 77.3 & -    & 71.6 &                                                                       & 38.5 & 50.9 & 69.5 & -    & 53.0 &                         & \multicolumn{1}{l}{42.8} & \multicolumn{1}{l}{48.8}                        & \multicolumn{1}{l}{67.4}                                               & \multicolumn{1}{l|}{\cellcolor[HTML]{C0C0C0}5.6}                        & \multicolumn{1}{l}{41.1} \\
\multirow{-8}{*}{\begin{tabular}[c]{@{}c@{}}VOC\\ 2007\end{tabular}}  & \multirow{-4}{*}{\begin{tabular}[c]{@{}c@{}}Faster\\ RCNN\\ (Retrain)\end{tabular}}        & 65.1            & 70.4 & 76.4 & 67.8 & 69.9 & \multirow{-4}{*}{\begin{tabular}[c]{@{}c@{}}CIFRCN\\ NP\end{tabular}} & 33.7 & 44.6 & 56.8 & 51.8 & 46.7 & \multirow{-4}{*}{Ours}  & \multicolumn{1}{l}{52.8} & \multicolumn{1}{l}{50.7}                        & \multicolumn{1}{l}{63.1}                                               & \multicolumn{1}{l|}{57.9}                                               & \multicolumn{1}{l}{56.1} \\ \hline
                                                                      &                                                                                            & 58.3            & -    & -    & -    & 58.3 &                                                                       & 58.1 & -    & -    & -    & 58.1 &                         & 49.2                     & -                                               & -                                                                      & -                                                                       & 49.2                     \\
                                                                      &                                                                                            & 33.5            & 21.0 & -    & -    & 27.3 &                                                                       & 39.0 & 24.6 & -    & -    & 31.8 &                         & 44.7                     & 15.7                                            & -                                                                      & -                                                                       & 30.2                     \\
                                                                      &                                                                                            & 20.1            & 11.9 & 26.4 & -    & 19.5 &                                                                       & 29.9 & 18.5 & 29.6 & -    & 26.0 &                         & 42.4                     & 13.1                                            & 14.0                                                                   & -                                                                       & 23.2                     \\
                                                                      & \multirow{-4}{*}{\begin{tabular}[c]{@{}c@{}}Faster\\ RCNN\\ (Fine-\\ tuning)\end{tabular}} & 16.8            & 9.3  & 9.4  & 31.4 & 16.7 & \multirow{-4}{*}{CIFRCN}                                              & 29.0 & 14.5 & 14.6 & 33.3 & 22.9 & \multirow{-4}{*}{IDWCF} & 38.0                     & 12.6                                            & 11.9                                                                   & 21.0                                                                    & 20.9                     \\ \cline{2-19} 
                                                                      &                                                                                            & 58.4            & -    & -    & -    & 58.4 &                                                                       & 55.6 & -    & -    & -    & 55.6 &                         & \multicolumn{1}{l}{46.5} & \multicolumn{1}{l}{\cellcolor[HTML]{C0C0C0}2.9} & \multicolumn{1}{l}{\cellcolor[HTML]{C0C0C0}0.1}                        & \multicolumn{1}{l|}{\cellcolor[HTML]{C0C0C0}0.2}                        & \multicolumn{1}{l}{12.4} \\
                                                                      &                                                                                            & 56.1            & 27.4 & -    & -    & 41.8 &                                                                       & 38.3 & 24.1 & -    & -    & 31.2 &                         & \multicolumn{1}{l}{18.8} & \multicolumn{1}{l}{16.9}                        & \multicolumn{1}{l}{\cellcolor[HTML]{C0C0C0}0.1}                        & \multicolumn{1}{l|}{\cellcolor[HTML]{C0C0C0}0.2}                        & \multicolumn{1}{l}{9.0}  \\
                                                                      &                                                                                            & 54.9            & 26.6 & 30.8 & -    & 37.4 &                                                                       & 26.2 & 18.6 & 27.2 & -    & 24.0 &                         & \multicolumn{1}{l}{10.8} & \multicolumn{1}{l}{10.1}                        & \multicolumn{1}{l}{20.3}                                               & \multicolumn{1}{l|}{\cellcolor[HTML]{C0C0C0}0.1}                        & \multicolumn{1}{l}{10.3} \\
\multirow{-8}{*}{\begin{tabular}[c]{@{}c@{}}COCO\\ 2017\end{tabular}} & \multirow{-4}{*}{\begin{tabular}[c]{@{}c@{}}Faster\\ RCNN\\ (Retrain)\end{tabular}}        & 54.8            & 26.3 & 28.0 & 37.2 & 36.6 & \multirow{-4}{*}{\begin{tabular}[c]{@{}c@{}}CIFRCN\\ NP\end{tabular}} & 26.5 & 15.1 & 15.5 & 32.2 & 22.3 & \multirow{-4}{*}{Ours}  & \multicolumn{1}{l}{38.4} & \multicolumn{1}{l}{15.9}                        & \multicolumn{1}{l}{12.3}                                               & \multicolumn{1}{l|}{26.9}                                               & \multicolumn{1}{l}{23.4} \\ \hline
\end{tabular}
\end{table*}

\noindent \textbf{Classes Split:} \label{sec:classes_split}
For class-incremental detection, we follow the classes split as \cite{hao2019ciod}.
As shown in Tab. \ref{tab:classsplit}, we split all classes into four class groups denoted by $\mathcal{G}_1$, $\mathcal{G}_2$, $\mathcal{G}_3$ and $\mathcal{G}_4$.
 Specifically, we sort all 20 classes of VOC 2007 alphabetically, split them into four class groups equally, so that each class group contains five classes. We remove images that contain objects from two or more class groups. As shown in Tab.~\ref{tab:classsplit}, we use $\mathcal{D}_{tr}(\cdot)$ and $\mathcal{D}_{te}(\cdot)$ to represent training and testing data that only contain objects of classes within the corresponding class group,  respectively.
 For MSCOCO 2017, we use the same split method as for VOC 2007, except that the class order in  MSCOCO 2017 is specified by itself. For the seen/unseen classes split for ZSD, we follow the split for VOC 2007 and 2012 proposed  by \cite{Demirel2018ZeroShotOD}. Particularly, \emph{car, dog, soft}, and \emph{train} as unseen classes, other classes as seen classes. There are two different types of seen and unseen classes splits setting for MSCOCO 2014: the 48/17 and 65/15 seen/unseen classes splits  proposed by \cite{bansal2018zero} and \cite{rahman2018polarity}, respectively. 

\begin{table}[]
   \centering
      \caption{Results of ZSD on COCO 2014 by 48/17 and 65/15 seen/unseen classes split}
   \label{tab:cocoresult}
      \small
\begin{tabular}{c|c|c}
\hline
Method & Seen/Unseen classes split & Unseen         \\ \hline
PLZSD  & 48/17             & 10.01          \\ \hline
Ours   & 48/17             & \textbf{15.29} \\ \hline
PLZSD  & 65/15             & 12.40          \\ \hline
Ours   & 65/15             & \textbf{20.67} \\ \hline
\end{tabular}
\end{table}

\noindent \textbf{Semantic Embedding:}
Many ZSD models use manual attributes and unsupervised word2vec/GloVe as semantic embeddings~\cite{rahman2018zero,zhu2019zero,fu2018recent}.
aPascal-aYahoo dataset~\cite{farhadi2009describing} has 64-dimensional binary attributes that characterize the visible objects for 20 object classes of Pascal VOC~\cite{fu2018recent}.
Therefore, we use the 64-dimensional attributes of aPascal-aYahoo dataset for Pascal VOC in our experiments.
Inspired by \cite{rahman2018polarity}, we use the $l_2$ normalized 300-dimensional unsupervised word2vec\cite{mikolov2013distributed} trained on  billions of words from texts like Wikipedia for MSCOCO in our experiments.

\begin{table}[]
\centering
  \caption{The result of ZSD on Pascal VOC 2007 val+test split}
\label{tab:zsdvoc2007}
\small
\begin{tabular}{c|c|c|c|c|c}
\hline
Method & car           & dog           & sofa          & train         & mAP        \\ \hline
HRM    & 55.0          & 82.0          & 55.0          & 26.0          & 54.5          \\ \hline
PLZSD  & \textbf{63.7} & 87.2          & 53.2          & \textbf{44.1} & \textbf{62.1} \\ \hline
Ours   & 47.26         & \textbf{89.3} & \textbf{57.0} & 41.4          & 58.7          \\ \hline
\end{tabular}
\end{table}

\noindent \textbf{Implementation:}
We use  Faster RCNN as the backbone. 
As  in Fig. \ref{fig:framework}, we use ResNet-101 as feature extractor that initialized by the weights pre-trained on ImageNet. 
Then we add  one FC layer for Pascal VOC and three FC layers for MSCOCO followed by three branches predicting class probability for seen and unseen classes and bounding box coordinates. 
We empirically set $\alpha=5$ in Eq. \ref{equ:mse}, margin $m=1$ in Eq. \ref{equ:tri} and $\beta=0.001$ in Eq. \ref{equ:lzsd}. $T$ in Eq. \ref{equ:inc} is set to 2 following the suggestion in \cite{hinton2015distilling}. In addition, we set $\gamma=2$ in Eq. \ref{equ:Lil} like  \cite{hao2019ciod}. To set an appropriate threshold $u_j$ of GPD for $j$-th class, we sort $d_{ij}$ in the descending order, then set $u_j = \eta len(d_{ij}), \eta=0.2$, where $len(\cdot)$ is the length of $d_{ij}$. We choose it as 80\% higher order quantile of the distances. We set $\delta=0.02$ in Eq. \ref{equ:pred} by grid search. We set the memory size $K=150$ and $K=600$ for Pascal VOC and MSCOCO, respectively. Our model is trained by 10 epochs for each incremental step with batch sizes of 3 images per GPU. We set the learning rate by 0.001 at the beginning of each incremental step, and decrease it by 0.2 after 5 epochs.

\subsection{Results on Incrementally Zero-Shot Detection}

We rigorously evaluate our IZSD-Ever on both VOC 2007 and COCO 2017 datasets. We train our model by incrementally adding new classes and semantic information. As shown in Tab. \ref{tab:classsplit}, in the first incremental step, we train the model on $\mathcal{D}_{tr}(\mathcal{G}_1)$ containing class group $\mathcal{G}_1$, and in the subsequent incremental steps, new classes are added continuously and trained using the corresponding train data. Tab. \ref{tab:IZSD_VOC} and Tab. \ref{tab:overallresult} present the  results on IZSD of our IZSD-Ever. As shown in Tab. \ref{tab:IZSD_VOC}, our model not only can keep the knowledge of the old class well, but also can effectively detect unseen classes. 
However, the performance of unseen classes is lower than that of seen classes, which is caused by the inherent domain drift problem \cite{fu2018recent} of ZSL. It refers to the difference in the data distribution of seen classes and unseen classes, which leads to performance degradation in unseen classes.
We have also observed that some unseen classes perform poorly. This is because the unseen class needs to be semantically similar to some seen classes to effectively transfer knowledge from the seen classes to the unseen classes recognition task.
Therefore, a good seen/unseen classes split can improve performance on unseen classes, as described in seen/unseen classes split in classes split \ref{sec:classes_split}.
In addition, in Tab. \ref{tab:overallresult}, the performance of our model  in COCO2017 is relatively worse than that in VOC2007. On the one hand, it is because COCO is more challenging than VOC2007. On the other hand, compared with seen/unseen classes split of MSCOCO, too many unseen classes result in poor performance on unseen classes.

\subsection{Results on Class-Incremental Detection}

\noindent \textbf{Compared Methods}: We compare against two state-of-the-art CIL detectors. (1) IDWCF~\cite{shmelkov2017incremental}. (2) CIFRCN~\cite{hao2019ciod}. (3) CIFRCN NP \cite{hao2019ciod} (a variant of CIFRCN). (4) Faster RCNN (fine-tuning) is trained only on data in new classes but no old classes. (5) Faster RCNN (retrain) is trained on both new classes data and old classes data, whose performance can be seen as an upper bound of our framework can achieve.

\noindent \textbf{Results}: 
We evaluate our IZSD-Ever and compared methods under the class-incremental detection setting. 
Tab. \ref{tab:overallresult} presents the results on class-incremental detection of our IZSD-Ever and compared methods on VOC 2007 and COCO 2017. 
Our model is better than the compared methods in most cases, not only can keep the knowledge of the old class well, but also can effectively detect unseen classes. However, the compared methods cannot detect unseen classes. In the fourth step, our model achieves high performance in four class groups for the reason that all classes are seen classes at this step. This also shows that the performance of the previous three steps of seen classes is not very high because the need to detect seen and unseen simultaneously. Therefore, IZSD is a very challenging task, and it is still far from totally addressing this task.

\subsection{Results on Zero-Shot Detection}
\noindent \textbf{Compared Methods}: Our model compares to two recent ZSD methods on zero-shot detection. (1) HRM \cite{Demirel2018ZeroShotOD} proposes a convex combination of embeddings used in conjunction with a detection network. (2) PLZSD \cite{rahman2018polarity} proposes a 'Polarity loss' to maximize the gap between positive and negative predictions. 

\noindent \textbf{Results}: 
We validate the performance of our model on ZSD tasks using VOC2007 +2012 and COCO 2014 datasets.
As shown in Tab. \ref{tab:cocoresult}, we present the results of our model and PLZSD on two different seen/unseen classes splits, and report mAP on the unseen classes for the zero-shot detection task. 
From the results, our model significantly outperforms PLZSD on the two different seen and unseen class splits on COCO 2014. For example, on the 48/17 seen/unseen classes split, the unseen result of our model reaches 15.29, while the results of PLZSD only 10.01. On the 65/15 seen/unseen classes split, our result reaches 20.67, while the PLZSD only reaches 12.40. 
As for the results of Pascal VOC, we compared our model with HRM and PLZSD, and the results are shown in Tab. \ref{tab:zsdvoc2007}. It can be seen from the results that our model has the highest AP on dog and sofa, and the results on car and train are comparable to the results on PLZSD. 
The mAP of our model exceeds the mAP of HRM and is comparable to the mAP of PLZSD.

\begin{table}[t]
\centering
   \caption{The results of the two baselines on Pascal VOC 2007}
\label{tab:baselineresult}
   \small
   \setlength{\tabcolsep}{0.12em}
\begin{tabular}{c|cccc|c|cccc|c}
\hline
\multirow{2}{*}{Step} & \multicolumn{5}{c|}{Baseline-$\mathcal{L}_{bone}$} & \multicolumn{5}{c}{Baseline-SI}   \\ \cline{2-11} 
                      & $\mathcal{G}_1$        & $\mathcal{G}_2$        & $\mathcal{G}_3$        & $\mathcal{G}_4$       & mAP     & $\mathcal{G}_1$     & $\mathcal{G}_2$    & $\mathcal{G}_3$    & $\mathcal{G}_4$    & mAP   \\ \hline
1                     & 24.17    & 1.92     & 1.78     & 0.32    & 7.05    & 49.80 &      &      &      & 49.80 \\ 
2                     & 10.59    & 34.99    & 3.65     & 3.55    & 13.20   & 43.35 & 3.84 &      &      & 23.59 \\ 
3                     & 10.78    & 19.14    & 42.12    & 2.46    & 18.62   & 35.74 & 4.23 & 3.23 &      & 14.40 \\ 
4                     & 8.23     & 16.64    & 23.36    & 29.08   & 19.33   & 34.76 & 3.45 & 4.42 & 2.95 & 11.40 \\ \hline
\end{tabular}
\end{table}

\subsection{Ablation Study}
\label{sec:ablation}

\begin{table}[t]
\centering
  \caption{Comparison with the method using general MSE loss on VOC 2007}
\label{tab:Lmse}
\small
\setlength{\tabcolsep}{0.12em}
\begin{tabular}{c|cccc|c|cccc|c}
\hline
\multirow{2}{*}{Step} & \multicolumn{5}{c|}{bfMSE} & \multicolumn{5}{c}{MSE}   \\ \cline{2-11} 
                      & $\mathcal{G}_1$        & $\mathcal{G}_2$        & $\mathcal{G}_3$        & $\mathcal{G}_4$       & mAP     & $\mathcal{G}_1$     & $\mathcal{G}_2$    & $\mathcal{G}_3$    & $\mathcal{G}_4$    & mAP   \\ \hline
1                     & 45.45                  & 1.65                   & 5.47                   & 2.10                   & 13.67                    & 38.88                  & 1.72                   & 4.15                   & 1.71                   & 11.61                    \\ 
2                     & 42.96                  & 53.42                  & 6.49                   & 5.22                   & 27.02                    & 28.12                  & 51.71                  & 5.98                   & 5.67                   & 22.87                    \\ 
3                     & 42.82                  & 48.78                  & 67.38                  & 5.59                   & 41.14                    & 26.86                  & 44.71                  & 58.31                  & 2.51                   & 33.10                    \\ 
4                     & 52.79                  & 50.73                  & 63.05                  & 57.86                  & 56.11                    & 42.99                  & 46.27                  & 52.56                  & 50.00                  & 47.96                    \\ \hline
\end{tabular}
\end{table}

\noindent \textbf{Compared with Two Baselines}: 
We compare our model with two baselines under the setting of incremental learning on VOC 2007. 
\textbf{Baseline-$\mathcal{L}_{bone}$}: the model is trained using $\mathcal{L}$ (Eq.~\ref{equ:Lbone}) along. 
\textbf{Baseline-SI}: the model that only adds the semantic information of the new(unseen) classes at from the second incremental step based on the data of the classes group $\mathcal{G}_1$, to recognize more and more new classes. 
Tab. \ref{tab:baselineresult} demonstrates that the performance of the Baseline-$\mathcal{L}_{bone}$ model in the old classes is significantly reduced, that is, catastrophic forgetting. The comparison between Tab. \ref{tab:baselineresult} and Tab. \ref{tab:overallresult} shows that $\mathcal{L}_{IL}$ can effectively alleviate catastrophic forgetting.
As shown in Tab. \ref{tab:baselineresult}, even though the Baseline-SI method can alleviate forgetting on the class group $\mathcal{G}_1$, there is a highly biased toward predicting the seen classes for the reason that the model is trained using data from the seen classes.
Compared with Tab. \ref{tab:overallresult}, the performance of our model on VOC 2007 on class group $\mathcal{G}_1$ is significantly better than that of Baseline-SI on class group $\mathcal{G}_1$. At the same time, the performance of our model in the unseen class is also better than that of Baseline-SI. 
Although Baseline-SI seems to be able to perform incremental learning using only the semantic information of the new classes, baseline-SI performs very poorly when there are more and more new classes.

\noindent \textbf{The Effect of bfMSE}: 
To demonstrate the effect of bfMSE loss (Eq. \ref{equ:mse}), we compare the results of the model using general MSE loss on VOC 2007. As shown in the Tab. \ref{tab:Lmse}, the performance of our bfMSE on both seen and unseen classes is better than general MSE loss. 
This shows that using our proposed bfMSE loss can effectively reduce the impact of this background and foreground imbalance.
More results are available in the supplementary material.

\section{Conclusion}
In this paper,  we propose an end-to-end model IZSD-Ever to solve IZSD by integrating ZSL and CIL in a single model. 
We also propose a novel extreme value analyzer to detect objects from old seen, new seen, and unseen classes simultaneously. Besides, we designed two novel losses, \emph{i.e.}, bfMSE loss and projection distance loss for zero-shot classifier. In our experiments, our model outperforms the alternative methods on Pascal VOC and MSCOCO datasets.

\bibliographystyle{IEEEtran}
% argument is your BibTeX string definitions and bibliography database(s)
\bibliography{IEEEabrv,mybibfile.bib}
%
% <OR> manually copy in the resultant .bbl file
% set second argument of \begin to the number of references
% (used to reserve space for the reference number labels box)
% \begin{thebibliography}{1}

% \bibitem{IEEEhowto:kopka}
% H.~Kopka and P.~W. Daly, \emph{A Guide to \LaTeX}, 3rd~ed.\hskip 1em plus
%   0.5em minus 0.4em\relax Harlow, England: Addison-Wesley, 1999.

% \end{thebibliography}

\section*{supplementary material}

   In this supplementary material, we provide more implementation details and experimental results about our proposed framework. We first describe the details of the four datasets used in our all experiments. Then we provide details of classes splits of different datasets in incremental learning and zero-shot detection. We also provide some qualitative results of our IZSD-EVer on VOC 2007. Finally, we provide more results.
   
\section*{More Experimental Details}
\subsection{Datasets}
In this section, we describe the details of the four datasets used in our experiments. As shown in Tab. \ref{tab:dataset}, the number of training data and testing data of VOC 2007 are 5,011 and 4,952, respectively. 
In order to make a fair comparison with other ZSD methods, we used the same dataset as them, \emph{i.e.}, VOC 2007+VOC 2012. Particularly, the training data of VOC 2007 + VOC 2012 is composed of train splits of  VOC 2007 and VOC 2012. The testing data of it consists of val+test split of VOC 2007. It can be seen that the amount of VOC 2007+VOC 2012 is significantly larger than VOC 2007, so the model can be better trained at VOC 2007+ VOC 2012 to achieve better performance for the unseen classes. As can be seen from Tab. \ref{tab:dataset}, the training data of COCO 2014 and COCO 2017 are 16 and 23 times that of VOC 2007, respectively.

\subsection{Classes splits}
In this section, we detail the classes splits of different datasets in incremental learning and zero-shot detection.
In order to compare with two state-of-the-art class incremental detectors IDWCF \cite{shmelkov2017incremental} and CIFRCN\cite{hao2019ciod}, we use the same settings as \cite{hao2019ciod}. As shown in Tab \ref{tab:classsplit}. We sorted the 20 classes of VOC 2007 alphabetically and then divided them into 4 class groups equally. As for COCO 2017, we divide all classes into 4 class groups according to the class order of the dataset itself. 
For the seen and unseen classes split of zero-shot detection, we follow the classes split method for VOC 2007 + VOC 2012 proposed in \cite{Demirel2018ZeroShotOD}. We also follow two different types of seen and unseen classes splits setting for MSCOCO 2014: the 48/17 and 65/15 seen and unseen splits proposed by \cite{bansal2018zero} and \cite{rahman2018polarity}, respectively.

\begin{table}[t]
   \caption{The size of the training data and testing data for different datasets used in our experiments, as well as the number of classes}
   \centering
      \setlength{\tabcolsep}{0.3em}
      \begin{tabular}{c|c|c|c}
      \hline
         Dataset & \# training data & \# testing data & \# Classes \\ \hline\hline
         VOC 2007 & 5,011 & 4,952 & 20 \\ \hline
         VOC 2007 + VOC 2012 & 8,218 & 7,462 & 20 \\ \hline
         COCO 2014 & 82,783 & 40,504 & 80 \\ \hline
         COCO 2017 & 118,287 & 5,000 & 80 \\ \hline
      \end{tabular}
   \label{tab:dataset}
\end{table}

\subsection{Implementation Details}
In this section, we will describe more implementation details.
We use Faster RCNN as the backbone. We follow \cite{he2016deep} that use those layers before conv5\_x of ResNet-101 as feature extractor, while conv5\_x is performed on RoI pooled feature map. 
The threshold $\delta$ used to determine whether the region proposal belongs to unseen is generally very small, and we set it to 0.02 in our experiments. 
We set the memory size K = 150 and K = 600 for Pascal VOC and MSCOCO respectively, because the number of old classes is 15 and 60 respectively in the third incremental step, so each class have 10 images. 
The proposed framework is implemented in PyTorch, and all experiments are performed on four NVIDIA GeForce GTX 1080Ti GPUs.

\section*{More Results}
\subsection{Qualitative Results}
\begin{figure*}[h]
    \centering
   \includegraphics[width=1.0\linewidth]{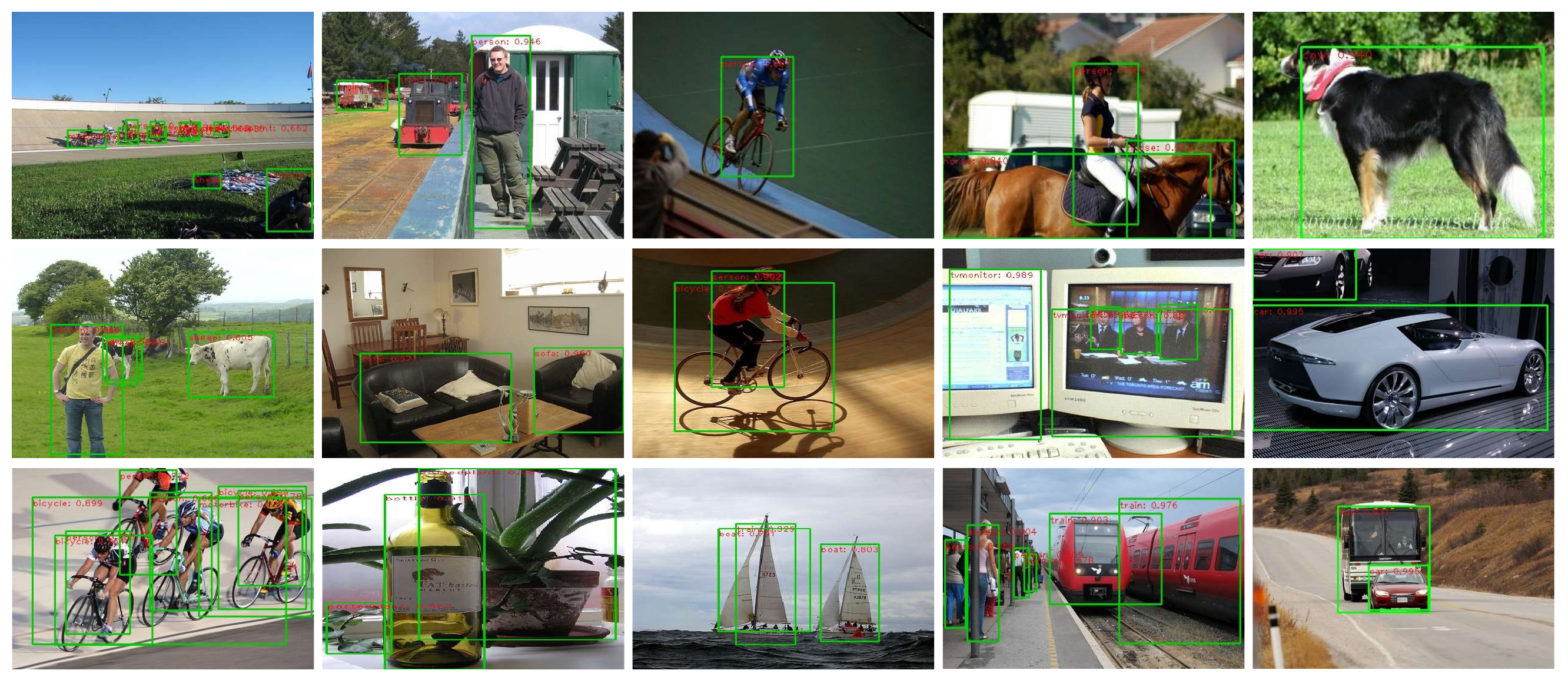}
      \caption{Qualitative Results of our IZSD-EVer on VOC 2007}
   \label{fig:qualitative_results}
\end{figure*}

As shown in Fig. \ref{fig:qualitative_results}, we provide some qualitative results of our IZSD-EVer on VOC 2007. 
% All visual results are generated by the IZSD-EVer model that can detect seen and unseen classes, simultaneously.

\subsection{The fitting of GPD} 
\begin{figure}[t]
  \centering
  \includegraphics[width=0.9\linewidth]{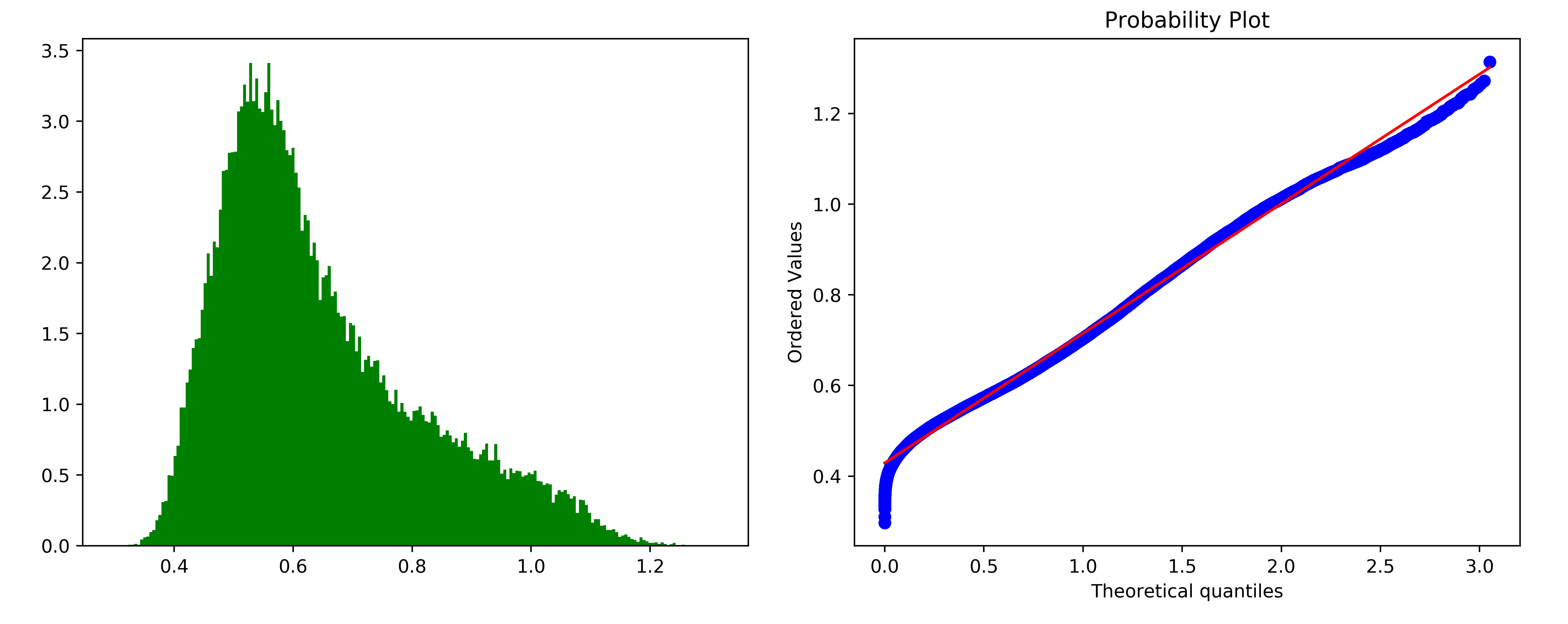}
      \caption{The histogram of the Euclidean distance from the projected semantic vector to the mean semantic vector for aeroplane class, and the Q-Q plot of the fitting  GPD.   \label{fig:histogram}}
\end{figure}

In EVer, we use the GPD to model the Euclidean distances from projected semantic vectors to the mean semantic vector for each seen classes. 
As shown in the left part of Fig. \ref{fig:histogram}, the histogram of the Euclidean distances presents a thick tail distribution, so EVT is well suited for modelling the Euclidean distances. To test the fitting of GPD, we used the most commonly used Quantile-Quantile (Q-Q) plot. Q-Q plot is a graphical technique for determining whether a certain two data sets are from the same distribution.

As can be seen from the right part of Fig. \ref{fig:histogram}, Q-Q plot is very approximate to a straight line, indicating that GPD fitting is very well. 

\begin{table}[t]
\centering
  \caption{Comparison with the method without projection distance loss on VOC 2007}
  \label{tab:Lpd}
     \small
  \setlength{\tabcolsep}{0.12em}
\begin{tabular}{c|cccc|c|cccc|c}
\hline
\multirow{2}{*}{Step} & \multicolumn{5}{c|}{$\mathcal{L}_{pd}^{zsc}$}                                 & \multicolumn{5}{c}{without $\mathcal{L}_{pd}^{zsc}$}                         \\ \cline{2-11} 
                      & $\mathcal{G}_1$ & $\mathcal{G}_2$ & $\mathcal{G}_3$ & $\mathcal{G}_4$ & mAP   & $\mathcal{G}_1$ & $\mathcal{G}_2$ & $\mathcal{G}_3$ & $\mathcal{G}_4$ & mAP   \\ \hline
1                     & 45.45           & 1.65            & 5.47            & 2.10            & 13.67 & 36.06           & 1.8             & 2.54            & 1.62            & 10.5  \\ 
2                     & 42.96           & 53.42           & 6.49            & 5.22            & 27.02 & 29.99           & 50.59           & 6.81            & 5.08            & 23.12 \\ 
3                     & 42.82           & 48.78           & 67.38           & 5.59            & 41.14 & 25.16           & 44.89           & 58.6            & 2.85            & 32.88 \\ 
4                     & 52.79           & 50.73           & 63.05           & 57.86           & 56.11 & 42.86           & 48.43           & 56.21           & 50.05           & 49.39 \\ \hline
\end{tabular}
\end{table}

\subsection{The effect of PD loss}
To demonstrate the effect of PD loss, we compare the results of the model without PD loss. As shown in the Tab. \ref{tab:Lpd}, the performance of our model on all classes is better than the model without PD loss. This shows that using our proposed PD loss can maintain the ability to align the visual space and
semantic spaces of old seen classes.

\subsection{The effect of $\alpha$ in $\mathcal{L}_{mse}$}
The hyperparameter $\alpha$ in $\mathcal{L}_{mse}$ is used to alleviate extreme imbalance between background and foreground in region proposals generated by RPN. To demonstrate the impact of $\alpha$, we fixed other hyperparameters to evaluate the performance of our IZSD-EVer with different $\alpha=\{1,3,5,7\}$. As shown in Fig. \ref{fig:ablation_alpha}, our IZSD-EVer with $\alpha=5$ performs the best for seen and unseen classes. Therefore, we set $\alpha=5$ in the main paper.

\subsection{The effect of $\beta$ in $\mathcal{L}_{zsc}$}
$\beta$ is a hyperparameter to control the importance of $\mathcal{L}_{rec}$ loss. We also do comparative experiments to evaluate the effect of $\beta$ in $\mathcal{L}_{zsc}$. 
We fixed other hyperparameters and only changed the  $\beta$ to $\{0.1, 0.01, 0.001,0.0001\}$. As shown in Fig. \ref{fig:ablation_beta}, our IZSD-EVer performs the best for seen and unseen classes where $\beta=0.001$.  Therefore, we set $\beta=0.001$ in the main paper.

\subsection{The effect of $\gamma$ in $\mathcal{L}_{IL}$}
The hyperparameter $\gamma$ controls the importance of $\mathcal{L}_{fd}^{rpn}$ in $\mathcal{L}_{IL}$. We do comparative experiments to evaluate the effect of $\gamma$. 
We fixed other hyperparameters and only changed the $\gamma$ to $\{1, 2, 5, 10\}$. As shown in Fig. \ref{fig:ablation_gamma}, our IZSD-EVer performs the best for seen and unseen classes where $\gamma=2$, which is consistent with the recommendation of Hao \emph{et al.}.  Therefore, we set $\gamma=2$ in the main paper.

\subsection{The effect of margin $m$ in $\mathcal{L}_{tri}$}
The hyperparameter $m$ is the margin in the $\mathcal{L}_{tri}$. To demonstrate the impact of margin $m$, we fixed other hyperparameters and only changed the margin $m$ to $\{1, 2, 5, 10\}$. As shown in the  Fig. \ref{fig:ablation_m}, as $m$ increases, the mAP of IZSD-EVer shows a slow downward trend. Our IZSD-EVer performs the best where $m=1$. Thus, we set $m=1$ in the main paper. 

\subsection{The effect of $\delta$ in EVer}
$\delta$ is the threshold used in EVer to distinguish whether the projected semantic vector comes from the seen classes or the unseen classes. To demonstrate the impact of $\delta$, we fixed other hyperparameters and only changed the $\delta$ to $\{0.01, 0.02, 0.05, 0.1, 0.2\}$. As shown in the  Fig. \ref{fig:ablation_delta}, our IZSD-EVer performs the best where $\delta=0.02$, so we set $\delta=0.02$ in the main paper.

\subsection{The effect of memory size $K$}
\begin{figure*}[t]
   \centering
      \subfigure[\small{The effect of $\alpha$}]{
         \includegraphics[width=0.29\linewidth]{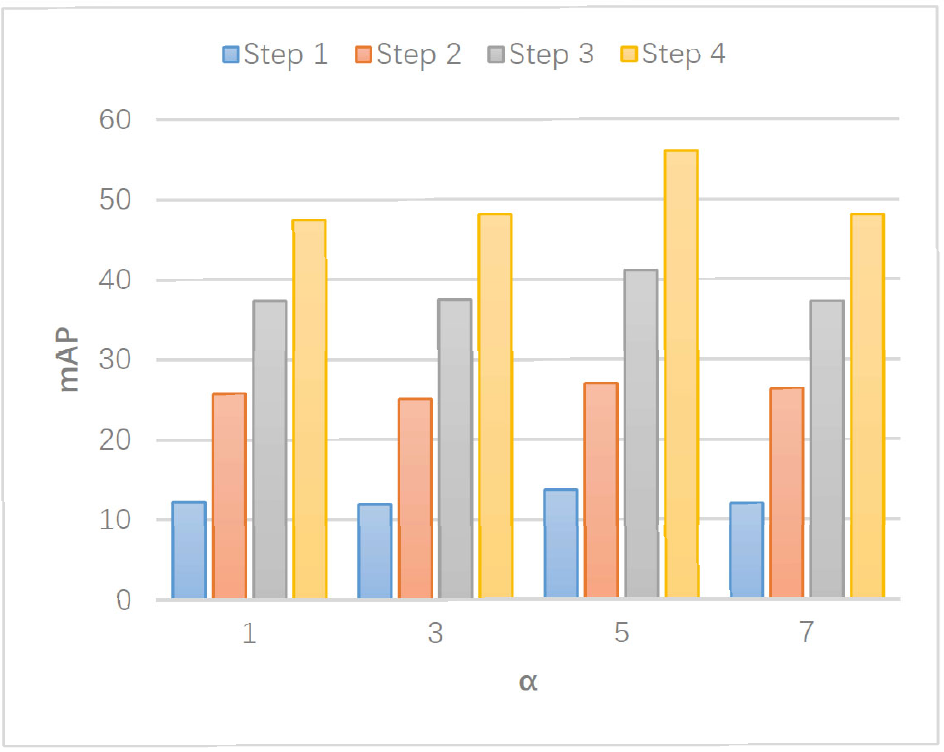}
         \label{fig:ablation_alpha}
      }
      \subfigure[\small{The effect of $\beta$}]{
         \includegraphics[width=0.29\linewidth]{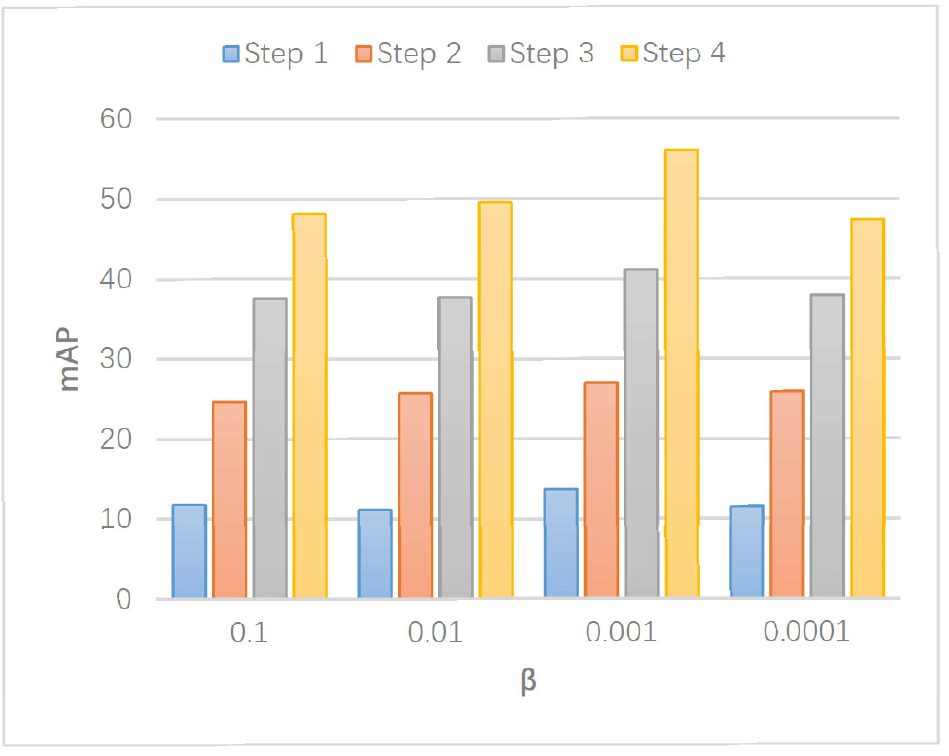}
         \label{fig:ablation_beta}
      }
      \subfigure[\small{The effect of $\gamma$}]{
         \includegraphics[width=0.29\linewidth]{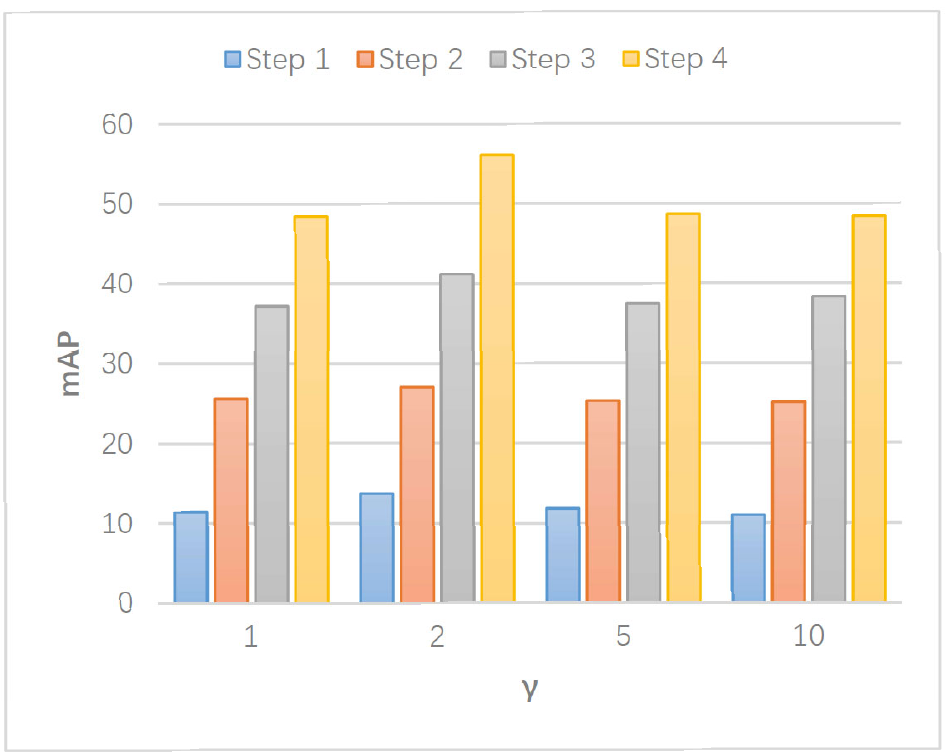}
         \label{fig:ablation_gamma}
      }
      \subfigure[\small{The effect of margin $m$}]{
         \includegraphics[width=0.29\linewidth]{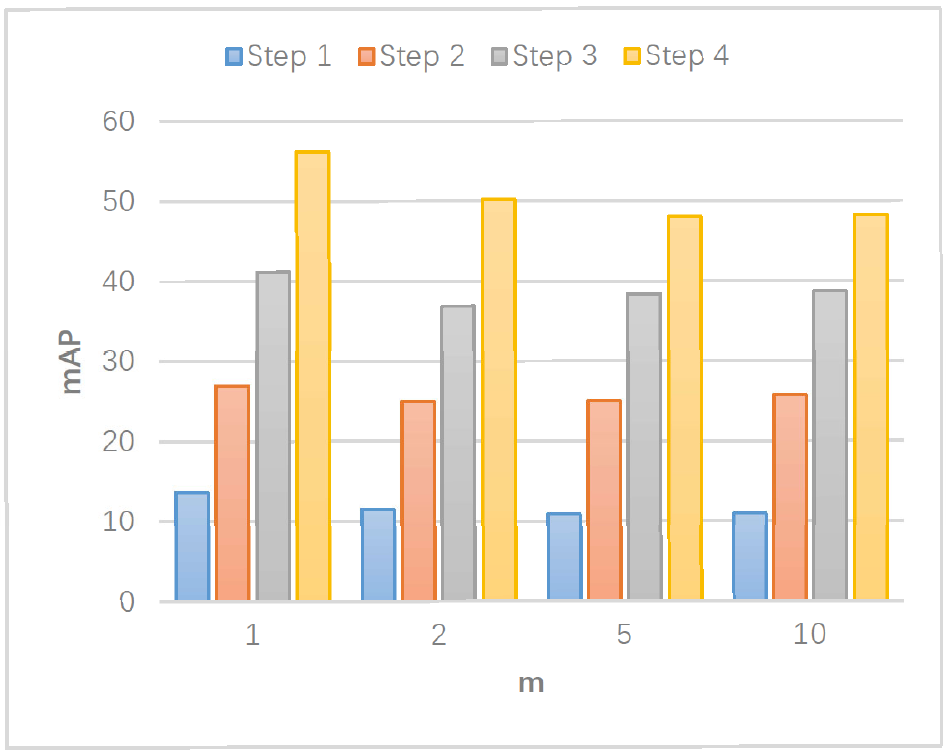}
         \label{fig:ablation_m}
      }
      \subfigure[\small{The effect of $\delta$}]{
         \includegraphics[width=0.29\linewidth]{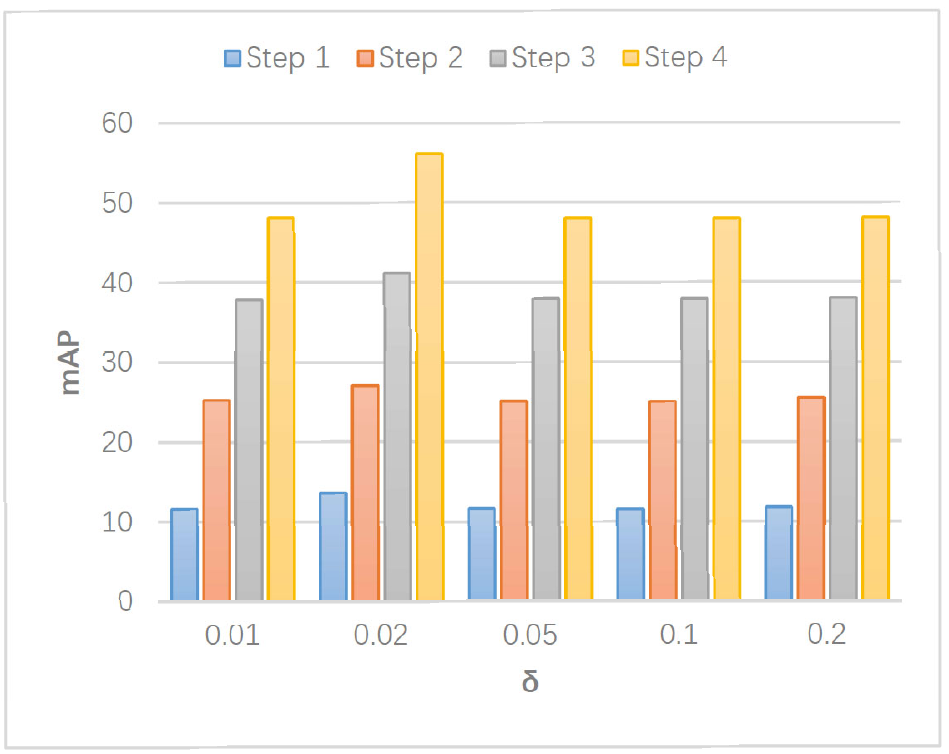}
         \label{fig:ablation_delta}
      }
      \subfigure[\small{The effect of memory size $K$}]{
         \includegraphics[width=0.29\linewidth]{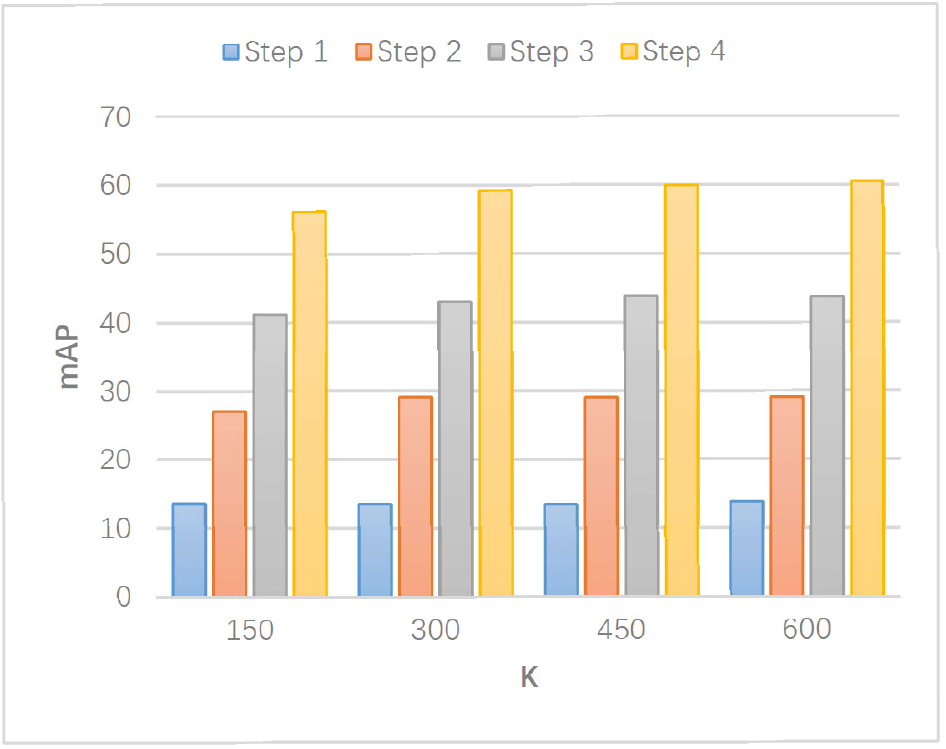}
         \label{fig:ablation_K}
      }
  \caption{The effect of different hyperparameters on the performance of IZSD-EVer.The horizontal axis of each subgraph is the different value of the hyperparameter, and the vertical axis is the mAP of each step.}
\end{figure*}

The bounded memory component $\mathcal{M}$ is introduced to alleviate the imbalance between old seen and new seen classes. 
To demonstrate the impact of memory size $K$, we fixed other hyperparameters to evaluate the performance of our IZSD-EVer with different $K=\{150,300,450,600\}$. As shown in the  Fig. \ref{fig:ablation_K}, as $K$ increases, the mAP of IZSD-EVer  gradually increases. However, the memory size cannot be too large, especially for devices with limited storage capacity, so we set $K=150$ in the main paper.

\begin{table*}[t]
    \centering
   \caption{Classes splits for different datasets in different experiments}
      \small
   \begin{tabular}{c|p{11em}|p{11em}|p{11em}|p{11em}}
      \hline
   Dataset & \multicolumn{4}{c}{\textbf{Classes Split for Incremental Learning}} \\ \hline
    & \multicolumn{1}{c}{$\mathcal{G}_1$} & \multicolumn{1}{c}{$\mathcal{G}_2$} & \multicolumn{1}{c}{$\mathcal{G}_3$} & \multicolumn{1}{c}{$\mathcal{G}_4$} \\ \hline
   VOC 2007 & aeroplane, bicycle, bird, boat, bottle & bus, car, cat, chair, cow & diningtable, dog, horse, motorbike, person & pottedplant, sheep, sofa, train, tvmonitor \\ \hline
   COCO 2017 & \begin{tabular}[c]{@{}p{10em}@{}}person, bicycle, car, motorcycle, airplane, bus, train, truck, boat, traffic light, fire hydrant, stop sign, parking meter, bench, bird, cat, dog, horse, sheep, cow\end{tabular} & \begin{tabular}[c]{@{}p{10em}@{}}elephant, bear, zebra, giraffe, backpack, umbrella, handbag, tie, suitcase, frisbee, skis, snowboard, sports ball, kite, baseball bat, baseball glove, skateboard, surfboard, tennis racket, bottle\end{tabular} & \begin{tabular}[c]{@{}p{10em}@{}}wine glass, cup, fork, knife, spoon, bowl, banana, apple, sandwich, orange, broccoli, carrot, hot dog, pizza, donut, cake, chair, couch, potted plant, bed\end{tabular} & \begin{tabular}[c]{@{}p{10em}@{}}dining table, toilet, tv, laptop, mouse, remote, keyboard, cell phone, microwave, oven, toaster, sink, refrigerator, book, clock, vase, scissors, teddy bear, hair drier, toothbrush\end{tabular} \\ \hline\hline
    & \multicolumn{4}{c}{\textbf{Classes Split for Zero-Shot Detection}} \\ \hline
    & \multicolumn{2}{c|}{\textbf{Seen Classes}} & \multicolumn{2}{c}{\textbf{Unseen Classes}} \\ \hline
   \begin{tabular}[c]{@{}c@{}}VOC 2007 + \\ VOC 2012\end{tabular}
     & \multicolumn{2}{l|}{\begin{tabular}[c]{@{}p{22em}@{}}bus, sheep, chair, person, boat, cat, motorbike, bottle, bird, aeroplane, diningtable, cow, bicycle, pottedplant, tvmonitor, horse\end{tabular}} & \multicolumn{2}{p{20em}}{car, dog, sofa, train} \\ \hline
   \begin{tabular}[c]{@{}c@{}}COCO 2014\\ 48/17\end{tabular} & \multicolumn{2}{l|}{\begin{tabular}[c]{@{}p{20em}@{}}toilet, bicycle, apple, train, laptop, carrot, motorcycle, oven, chair, mouse, boat, kite, sheep, horse, sandwich, clock, tv, backpack, toaster, bowl, microwave, bench, book, orange, bird, pizza, fork, frisbee, bear, vase, toothbrush, spoon, giraffe, handbag, broccoli, refrigerator, remote, surfboard, car, bed,  banana, donut, skis, person, truck, bottle, suitcase, zebra\end{tabular}} & \multicolumn{2}{l}{\begin{tabular}[c]{@{}p{22em}@{}}umbrella, cow, cup, bus, keyboard, skateboard, dog, couch, tie, snowboard, \\ sink, elephant, cake, scissors, airplane, cat, knife\end{tabular}} \\ \hline
   \begin{tabular}[c]{@{}c@{}}COCO 2014\\ 65/15\end{tabular} & \multicolumn{2}{l|}{\begin{tabular}[c]{@{}p{22em}@{}}person, bicycle, car, motorcycle, bus, truck, boat, traffic light, fire hydrant, stop sign, bench, bird, dog, horse, sheep, cow, elephant, zebra, giraffe, backpack, umbrella, handbag, tie, skis, sports ball, kite, baseball bat, baseball glove, skateboard, surfboard, tennis racket, bottle, wine glass, cup, knife, spoon, bowl, banana, apple, orange, broccoli, carrot, pizza, donut, cake, chair, couch, potted plant, bed, dining table, tv, laptop, remote, keyboard, cell phone, microwave, oven, sink, refrigerator, book, clock, vase, scissors, teddy bear, toothbrush\end{tabular}} & \multicolumn{2}{l}{\begin{tabular}[c]{@{}p{22em}@{}}airplane, train, parking meter, cat, bear, suitcase, frisbee, snowboard, fork, sandwich, hot dog, toilet, mouse, toaster, hair drier\end{tabular}} \\ \hline
   \end{tabular}
   \label{tab:classsplit}
\end{table*}

% that's all folks
\end{document}